\documentclass[10pt]{article} 

\usepackage[colorlinks,urlcolor=blue,linkcolor=blue,citecolor=blue]{hyperref}

\usepackage{fancyhdr}
\usepackage{titlesec}
\usepackage{natbib}

\usepackage{bm}
\usepackage{bbm}
\usepackage{bbold}
\usepackage{booktabs}
\usepackage{graphicx, wrapfig} 
\usepackage{amsmath}
\usepackage{amsthm} 
\usepackage{amssymb}
\usepackage{multirow}
\usepackage{multicol}   
\usepackage{caption}
\usepackage{subcaption}
\usepackage{wasysym}

\usepackage{enumitem}
\usepackage{arydshln}
\usepackage[bottom]{footmisc}
\usepackage{wrapfig}
\usepackage{mathtools}
\usepackage{dsfont}
\usepackage{makecell}

\usepackage{color, colortbl, xcolor}
\definecolor{Gray}{gray}{0.9}

\def\arrvline{\hfil\kern\arraycolsep\vline\kern-\arraycolsep\hfilneg}

\PassOptionsToPackage{hyphens}{url}\usepackage{hyperref}
\usepackage{xurl}

\usepackage[font=small]{caption}

\titleformat*{\section}{\large\bfseries}
\renewcommand{\headrulewidth}{0.2pt}

\begin{document}

\title{Finding the right XAI method --- A Guide for the Evaluation and Ranking of Explainable AI Methods in Climate Science}

\author{
 \normalsize{Philine Bommer$^{1,2,\dagger}$}
\qquad
 \normalsize{Marlene Kretschmer$^{3,4}$}
\qquad
 Anna Hedström$^{1,2}$
\\
\qquad
 Dilyara Bareeva$^{1,2}$
\qquad
 Marina M.-C. Höhne$^{2,5,6,7,\dagger}$
\vspace{5mm}
\\
{\small
\textit{$^1$ Department of Machine Learning,
Technische Universität Berlin,
10587 Berlin,
Germany}}
\vspace{-1mm}
\\
{\small
\textit{$^2$ Understandable Machine Intelligence Lab, Department of Data Science,
ATB, 14469  Potsdam, Germany}
}
\vspace{-1mm}
\\
{\small
\textit{$^3$ Institute for Meteorology, University of  Leipzig, Leipzig, German}}
\vspace{-1mm}
\\
{\small
\textit{$^4$ Department of Meteorology, University of Reading, Reading, UK}}
\vspace{-1mm}
\\
{\small
\textit{$^5$ BIFOLD – Berlin Institute for the Foundations of Learning and Data,
10587 Berlin,
Germany}
}
\vspace{-1mm}
\\
{\small
\textit{$^6$ Machine Learning Group, UiT the Arctic University of Norway, 9037 Tromsø, Norway}
}
\vspace{-1mm}
\\
{\small
\textit{$^7$ Department of Computer Science,
University of Potsdam, 14476 Potsdam, Germany}
}
\vspace{-1mm}
\\
{\small
\textit{$^\dagger$ corresponding authors}
}
}

\date{\vspace{-5ex}}


\newcommand{\fix}{\marginpar{FIX}}
\newcommand{\new}{\marginpar{NEW}}

\def\month{MM}  
\def\year{YYYY} 

\maketitle

\begin{abstract}
Explainable artificial intelligence (XAI) methods shed light on the predictions of machine learning algorithms. Several different approaches exist and have already been applied in climate science. However, usually missing ground truth explanations complicate their evaluation and comparison, subsequently impeding the choice of the XAI method. 
Therefore, in this work, we introduce XAI evaluation in the climate context and discuss different desired explanation properties, namely robustness, faithfulness, randomization, complexity, and localization. To this end, we chose previous work as a case study where the decade of annual-mean temperature maps is predicted. After training both a multi-layer perceptron (MLP) and a convolutional neural network (CNN), multiple XAI methods are applied and their skill scores in reference to a random uniform explanation are calculated for each property. Independent of the network, we find that XAI methods Integrated Gradients, layer-wise relevance propagation, and input times gradients exhibit considerable robustness, faithfulness, and complexity while sacrificing randomization performance. Sensitivity methods -- gradient, SmoothGrad, NoiseGrad, and FusionGrad, match the robustness skill but sacrifice faithfulness and complexity for randomization skill. We find architecture-dependent performance differences regarding robustness, complexity and localization skills of different XAI methods, highlighting the necessity for research task-specific evaluation. Overall, our work offers an overview of different evaluation properties in the climate science context and shows how to compare and benchmark different explanation methods, assessing their suitability based on strengths and weaknesses, for the specific research problem at hand. By that, we aim to support climate researchers in the selection of a suitable XAI method.
\end{abstract} %

\section{Introduction}

Deep learning (DL) has become a widely used tool in climate science and assists various tasks, such as nowcasting \citep{Xingjian15, Han_2017, Bromberg19}, climate or weather monitoring \citep{Hengl2017, Anant19} and forecasting \citep{Ham2019,Chen2020,Scher2021}, numerical model enhancement \citep{Yuval_2020,Harder21}, and up-sampling of satellite data \citep{Wang2021,Leinonen2021}. However, a deep neural network (DNN) is mostly considered a black box due to its inaccessible decision-making process. This lack of interpretability limits their trustworthiness and application in climate research, as DNNs should not only have high predictive performance but also provide accessible and consistent predictive reasoning aligned with existing theory \citep{McGovern2019,mamalakis2020explainable, CampsValls2020, Sonnewald2021, Clare2022, flora2022}.
Explainable artificial intelligence (XAI) aims to address the lack of interpretability by explaining potential reasons behind the predictions of a network. In the climate context, XAI can help to validate DNNs and on a well-performing model provide researchers with new insights into physical processes \citep{EbertUphoff2020, Hilburn2021}. For example, \citet{Gibson_2021} demonstrated using XAI that DNNs produce skillful seasonal precipitation forecasts based on known relevant physical processes. Moreover, XAI was used to improve the forecasting of droughts \citep{Dikshit_2021}, teleconnections \citep{Mayer2021}, and regional precipitation \citep{Pegion2022}, to assess external drivers of global climate change \citep{Labe_2021} and to understand sub-seasonal drivers of high-temperature summers \citep{van_Straaten_2022}. Additionally, \citet{Labe2022} showed that XAI applications can aid in the comparative assessment of climate models.\\
\begin{table}[t!]
  \caption{
  Overview and categorization of research on the transparency and understandability of neural networks. For this categorization we follow works like \citet{samek2019explainable, Ancona2019, Mamalakis2021a, Letzgus2021, flora2022}}
  \centering
  \resizebox{\textwidth}{!}{
    \begin{tabular}{|p{2cm}|p{4cm}|p{7.9cm}|}
    \hline
    \hline
   
    & &\textbf{local} (e.g. Shapley values \citep{Lundberg2017} or LRP \citep{Bach2015})\\ 
    \cline{3-3}
    &explanation target&\textbf{global} (e.g. activation-maximization \citep{Simonyan2014} or DORA \citep{bykov2022dora})\\
    \cline{2-3}
    \multirow{5}{*}{\textit{post-hoc}}&& \textbf{model-aware} (e.g. gradient \citep{Baehrens2010} ,LRP \citep{Montavon2019} or GradCAM \citep{selvaraju2017grad})\\
    \cline{3-3}
    &components&\textbf{model-agnostic} (e.g. LIME \citep{Ribeiro2016} or Shapley values \citep{Lundberg2017})\\
    \cline{2-3}
    &&\textbf{sensitivity} (e.g. gradient \citep{Baehrens2010} and GradCAM \citep{selvaraju2017grad})\\
    &explanation output& \textbf{feature contribution}|  \textbf{salience} |\textbf{attribution}  (e.g. Integrated Gradients \citep{sundararajan2017axiomatic} or LRP \citep{Bach2015})\\
    && \textbf{examples} (e.g. RISE \citep{petsiuk2018rise})\\
    \cline{1-3}
    \multirow{3}{*}{\textit{ante-hoc}}&& prototype network \citep{protopnet,gautam2022protovae,gautam2023looks}\\
    &self-explaining network& concept networks \citep{alvarez2018towards}\\
    && contrastive networks \citep{sawada2022c}\\
    \hline
    \hline
    \end{tabular}}
\label{tab:XAI}
\end{table}
Generally, explainability methods can be divided into ante-hoc and post-hoc approaches \citep{samek2019explainable} (see Table \ref{tab:XAI}). Ante-hoc approaches modify the DNN architecture to improve interpretability, like adding an interpretable prototype layer to learn humanly understandable representations for different classes (see e.g. \citet{protopnet} and \citet{gautam2022protovae, gautam2023looks}) or constructing mathematically similar but interpretable models \citep{Hilburn2023}. Such approaches are also called self-explaining neural networks and link to the field of interpretability \citep{flora2022}.
Post-hoc XAI methods, on the other hand, can be applied to any neural network architecture \citep{samek2019explainable} and here we focus on three characterizing aspects \citep{samek2019explainable, Letzgus2021, Mamalakis2021a}, as shown in Table \ref{tab:XAI}. The first considers the explanation target (i.e. what is explained) which can differ between local and global decision-making. While local explanations provide explanations of the network decision for a single data point \citep{Baehrens2010,Bach2015,vidovic2016feature,Ribeiro2016}, e.g., by assessing the contribution of each pixel in a given image based on the predicted class, global explanations reveal the overall decision strategy, e.g. by showing a map of important features or image patterns, learned by the model for the whole class \citep{Simonyan2014,vidovic2015opening, Nguyen2016, Lapuschkin2019, grinwald2022visualizing, bykov2022dora}. 
The second aspect concerns the components used to calculate the explanation, differentiating between model-aware and model-agnostic methods. Model-aware methods use components of the trained model for the explanation calculation, such as network weights, while model-agnostic methods consider the model as a black box and only assess the change in the output caused by a perturbation in the input \citep{strumbelj2010efficient,Ribeiro2016}. 
The third aspect considers the DNN explanation output. Here we can differentiate between methods where the assigned value of a pixel indicates the sensitivity of the network regarding that pixel also called \textit{sensitivity methods}, such as absolute gradient, as well as methods, that display the positive or negative contribution of a pixel to predict the class, such as layer-wise Relevance Propagation (see Section \ref{sec:XAI}) also called \textit{salience methods}, and methods presenting input examples leading to the same prediction. Beyond these three characteristics, recent efforts \citep{flora2022} also differentiate between feature importance methods encompassing mostly global methods, which calculate feature contribution based on the network performance (e.g. accuracy), and feature relevance methods describing mostly local methods which calculate contributions to the model prediction.
In climate research, the decision patterns learned by DNNs have been analyzed with local explanation methods such as LRP or Shapley values \citep{Gibson_2021,Dikshit_2021,Mayer2021, Labe_2021, he2021sub, felsche2021applying, Labe2022}. 
However, different local explanation methods can identify different input features as being important to the network decision, subsequently leading to different scientific conclusions \citep{Lundberg2017, han2022, flora2022}. Thus, with the increasing number of XAI methods available, selecting the most suitable method for a specific task poses a challenge and the practitioner's choice of a method is often based upon popularity or upon easy-access \citep{krishna2022}. To navigate the field of XAI, recent climate science publications have compared and assessed different explanation techniques using benchmark datasets, where the XAI method was assessed by comparing its predictions with a defined target, considered as ground truth \citep{Mamalakis2021a, Mamalakis_2022}. While benchmark datasets \citep{Yang2019, Arras2020, agarwal2022openxai} certainly contribute to the understanding of local XAI methods, the existence of a ground truth explanation is highly debated (e.g., \citet{Janzing2019, Sturmfels2020}). In the case of DNNs, ground truth explanation labels can only be considered approximations and are not guaranteed to align precisely with the model's decision process or the features it utilizes \citep{Ancona2019,  metaquantus2023}. For exact ground truth, either perfect knowledge of how the model handles the available information or a carefully engineered model would be required, which is usually not the case. Additionally, post-hoc explanation methods are generally only approximations of a model's behavior \citep{Lundberg2017, han2022}, and the distinct mathematical concepts of the different methods would consequently lead to distinct ground truth explanations. 

\begin{figure}[t!]
        \centering
        \includegraphics[width=0.85\textwidth]{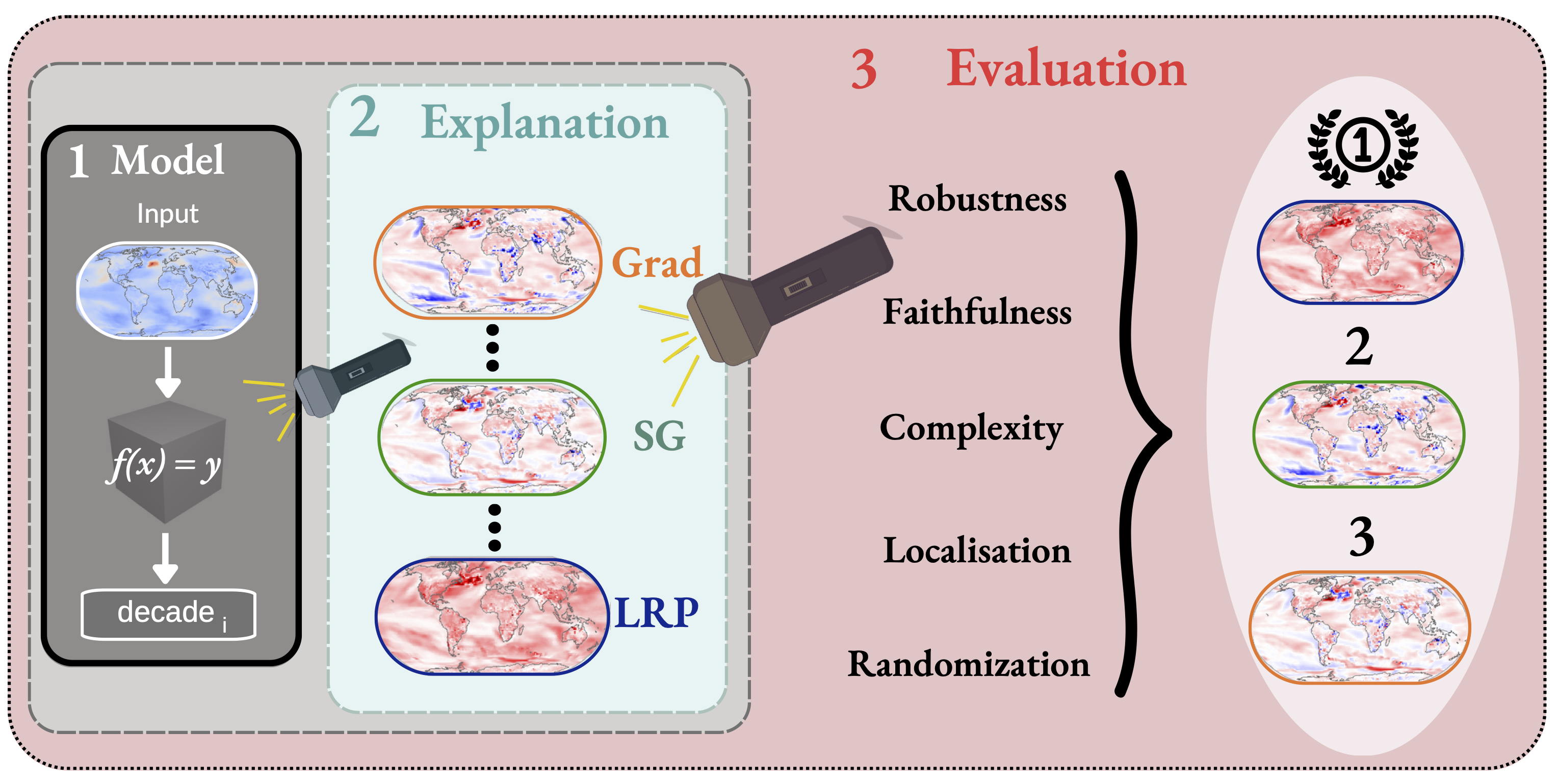}
        \caption{Schematic of the XAI evaluation procedure. Based on an annual temperature anomaly map as input, the network predicts the respective decade (box $1$). The explanation methods (Grad - gradient, SG - SmoothGrad applied to gradient, LRP - layer-wise relevance propagation) then provide insights (i.e., "shine a light", see box $2$) into the specific network's decision. The different explanation maps (marked in orange - Grad, green - SG,  and blue - LRP) highlight different areas as positively (red) and negatively (blue) contributing to the network decision. Here XAI evaluation can 'shine a light' on the explanation methods and help choose a suitable method (here indicated by the first rank) since evaluation explores the explanation maps regarding their robustness, faithfulness, localization, complexity, and randomization properties.} 
\label{fig:firstGraph}
\end{figure}

Here, we address these challenges, by introducing \textit{XAI evaluation} in the context of climate science to compare different local explanation methods. The field of XAI evaluation has emerged recently and refers to the development of metrics to compare, benchmark, and rank explanation methods, in different explainability contexts (e.g. \citet{Adebayo2018, Hedstroem22, metaquantus2023}). As discussed below in more detail, using evaluation metrics we are able to quantitatively assess the robustness, complexity, localization, randomization, and faithfulness of explanation methods, making them comparable regarding their suitability, their strengths, and weaknesses \citep{Hoffman2018, Barredo_Arrieta_2020, Mohseni2021, Hedstroem22}.\\
 
In this work, we discuss these properties in an exemplary manner and build upon work from \citep{Labe_2021}. In their work, an MLP was trained with global annual temperature anomaly maps and the network´s task was to assign the respective year or decade of occurrence. The MLP achieves the assignment, as global-mean warming progresses. Using layer-wise relevance propagation (LRP) they then identified the signals relevant to the network´s decision and found the North Atlantic, Southern Ocean, and Southeast Asia as key regions. Here, we use their work as a case study and train an MLP as well as a CNN for the same prediction tasks (see step 1 in Fig. \ref{fig:firstGraph}). Then, we apply several explanation methods and show the variation in their explanation maps, potentially leading to different scientific insights (step 2 in Figure \ref{fig:firstGraph}). We therefore introduce XAI evaluation metrics and quantify the skill of the different XAI methods against a random baseline in different properties to compare their performance with respect to the underlying task. 

This paper is structured as follows. In Section \ref{sec:background} we discuss the used dataset and network types, and briefly describe the different analysed explanation methods. Section \ref{sec:quantus} introduces XAI evaluation and describe five evaluation properties. 
Then, in Section 4, we first discuss the performance of both network types and provide a motivational example highlighting the risks of an uninformed choice of an explanation method. Next, we evaluate different XAI methods applied to the MLP, using two different metrics for each evaluation property, and then compare the XAI evaluation results for the different networks (see Section \ref{sec:metrics} and \ref{sec:Comparison}). Finally, in Section \ref{sec:choice}, we present a guideline on using XAI evaluation to choose a suitable XAI method. The discussion of our results and our conclusion are presented in Section \ref{sec:discuss}.

\section{Data and Methods}\label{sec:background}

\subsection{Data}\label{sec:Data}

We analyze data simulated by the general climate model, CESM1 \citep{Hurrell_2013}, focusing on the ``ALL’’ configuration \citep{Kay2015}, which is discussed in detail in \citet{Labe_2021}.
We use the global $2$-m air temperature (T2m) maps from $1920$ to $2080$. The data $\bm{\Omega}$ consist of $\mathrm{I} =40$ ensemble members $\Omega_{i\in\{1, \dots, I\}}$, and each member is generated by varying the atmospheric initial conditions $z_i$ with fixed external forcing $\theta_{\mathrm{clima}}$. Following \citet{Labe_2021}, we compute annual averages and apply a bilinear interpolation. This results in $T=161$ temperature maps for each member $\Omega_i\in \mathbb{R}^{T \times v\times h}$, with $v=144$ and $h=95$ denoting the number of longitudes and latitudes, with $1.9^{\circ}$ sampling in latitude and $2.5^{\circ}$ sampling in longitude. Accordingly, the whole dataset $\mathbf{X} \in \mathbb{R}^{\mathrm{I} \times T \times v\times h}$ contains $\mathrm{I} \times T$ samples. The data is split into a training $\Omega_{\mathrm{tr}}$ and a test set $\Omega_{\mathrm{test}}$. More precisely, we sample $20\%$ of the ensemble members (i.e., in total 8 ensemble members) as a test set $\mathbf{X}_{\text{test}}\in \mathbb{R}^{0.2\mathrm{I}\times T \times v\times h}$, and use the remaining $80\%$ (i.e., $32$ ensemble members) for training and validation. Of these $32$ ensemble members all temperature maps are split into a training ($80\%$ of the data points, i.e. $64\%$ of all ensemble members) and validation ($20\%$ of the temperature maps, i.e.$16\%$ of all ensemble members) set. All temperature maps $x \in \mathbb{R}^{vxh}$ are standardized by subtracting the mean and subsequently dividing by the corresponding standard deviation at each grid-point individually, whereby the mean $\mathbf{x}_{\text{mean}}\in \mathbb{R}^{v\times h}$ and standard deviation $\mathbf{x}_{\text{std}}\in \mathbb{R}^{v\times h}$ are computed over the training set only.

\subsection{Networks}\label{sec:network}
Following \citet{Labe_2021}, we train an MLP, $f_{\text{MLP}}:\mathbb{R}^d \rightarrow \mathbb{R}^c$ with network weights $W\in\mathcal{W}$, to solve a fuzzy classification problem by combining classification and regression. As input $\mathbf{x} \in \Omega$, the network considers the flattened temperature maps with dimensionality $d = v\times h$. Given the goal of fuzzy classification, first, the network assigns each map to one of the $C = 20$ different classes, where each class corresponds to one decade between $1900$ and $2100$ (see Figure 1 in \citet{Labe_2021}). The network output $f(x)$, thus, is a probability vector $\mathbf{y}\in \mathbb{R}^{1\times C}$ across $C = 20$ classes. Afterward, since the network can assign a nonzero probability to more than one class, regression is used to predict the year $\hat{y}$ of the input as:
\begin{equation}
    \hat{y} = \sum_{i=1}^{C}y_i  \bar{Y}_i,
    \label{eq:reg}
\end{equation}
where $y_i$ is the probability of class $i$, predicted by the network $\mathbf{y}=f(\mathbf{x})$ in the classification step, and $\bar{Y}_i$ denotes the central year of the corresponding decade class $i$ (e.g. for class $i=1$,  $\bar{Y}_1=1925$ represents the decade $1920-1929$). Accordingly, the task ensures the association of temperature patterns to the respective year or decade. Here we train using the binary cross-entropy loss, considering Eq. \eqref{eq:reg} only for performance evaluation.

Additionally, we construct a CNN $f_{\text{CNN}}:\mathbb{R}^{v\times h} \rightarrow \mathbb{R}^c$ that maintains the longitude-latitude grid of the data $\mathbf{x}_{\text{img}}\in\mathbb{R}^{v\times h}$ for each input sample (see Section \ref{sec:Data}), unlike the flattened input used for the MLP. The CNN consists of a 2D-convolutional layer (2dConv) with $6\times6$ window size and a stride of 2. The second layer includes a max-pooling layer with a $2\times2$ window size, followed by a dense layer with $L^2$-regularization and a softmax output layer.

\subsection{Explainable Artificial Intelligence (XAI)}\label{sec:XAI}
In this work, we focus on local model-aware explanation methods belonging to the group of feature-attribution methods \citep{Ancona2019,Das2020,zhou2022}. 
For the mathematical details, we refer to Appendix \ref{app:a1}.

\paragraph{Gradient \citep{Baehrens2010}}explains the network decision by computing the first partial derivative of the network output $f(\mathbf{x})$ with respect to the input. This explanation method feeds backward the network’s prediction to the features in the input $\mathbf{x}$, indicating the change in network prediction given a change in the respective features. The explanation values correspond to the network's \textit{sensitivity} to each feature, thus belonging to the group of sensitivity methods. The absolute gradient, often referred to as Saliency map, can also be used as an explanation \citep{Simonyan2014}.

\paragraph{Input times gradient}is an extension of the gradient method and computes the product of the gradient and the input. In the explanation map, a high relevance is assigned to an input feature if it has a high value and the model gradient is sensitive to it. Therefore, contrary to the gradient as a sensitivity method, input times gradient and other methods including the input pixel value are considered salience methods \citet{Ancona2019} (or attribution methods, e.g. \citet{Mamalakis_2022}). 

\paragraph{Integrated Gradients \citep{sundararajan2017axiomatic}}extends input times gradient, by integrating a gradient along a line path from a baseline (generally a reference vector for which the network’s output is zero, e.g. all zeros for standardized data) to the explained sample $\mathbf{x}$. In practice, the gradient explanations of a set of images lying between the baseline and $\mathbf{x}$ are averaged and multiplied by the difference between the baseline and the explained input (see Eq. \eqref{integratedgrad}). Hence, the Integrated Gradients method is a salience method and highlights the difference between the features important to the prediction of $\mathbf{x}$ and features important to the prediction of the baseline value.

\paragraph{Layerwise Relevance Propagation(LRP) \citep{Bach2015, Montavon2019}} computes the relevance for each input feature by feeding the network's prediction backward through the model, layer by layer, until the prediction score is distributed over the input features and is a salience method. Different propagation rules can be used, all resembling the energy conservation rule, i.e., the sum of all relevances within one layer is equal to the original prediction score.
In case of the \textbf{$\bm{\alpha}$-$\bm{\beta}$-rule}  
relevance is assigned at each layer to each neuron. All positively contributing activations of connected neurons in the previous layer are weighted by $\alpha$, while $\beta$ is used to weight the contribution of the negative activations. The default values are $\alpha = 1$ and $\beta=0$, where only positively contributing activations are considered. Contrary to that, the \textbf{z-rule} calculates the explanation by including both negative and positive neuron activations. Hence, the corresponding explanations, visualized as heatmaps, display both positive and negative evidence. The \textbf{composite rule} combines various rules for different  layer types. The method accounts for layer structure variety in CNNs, such as fully connected, convolutional, and pooling layers.

\paragraph{SmoothGrad \citep{smilkov2017}}aims to filter out the background noise (i.e., the gradient shattering effect, where gradients resemble white noise with increasing layer number \citep{Balduzzi2017}) to enhance the interpretability of the explanation. To this end, multiple noisy samples are generated by adding random noise to the input, then the explanations of the noisy samples are computed and averaged, such that the most important features are enhanced and the less important features are "canceled out".

\paragraph{NoiseGrad \citep{Bykov2021a}}perturbs the weights of the model, instead of the input feature as done by SmoothGrad. The explanations, resulting from explaining the predictions made by the noisy versions of the model on the same image, are averaged to suppress the background noise of the image in the final explanation.

\paragraph{FusionGrad \citep{Bykov2021a}}combines SmoothGrad and NoiseGrad by perturbing both the input features and the network weights. The purpose of the method is to account for uncertainties within the network and the input space \citep{Bykov2022}.

\paragraph{Deep SHapley Additive exPlanations (DeepSHAP) \citep{Lundberg2017}}estimates Shapley values for the full DNN by dividing it into small network components, calculating the Shapley values, and averaging them across all components. The idea behind SHAP (SHapley Additive exPlanations) values is to fairly distribute the contribution of each feature to the prediction of a specific instance considering all possible feature combinations. Following the game-theoretic concept of Shapley values \citep{shapley1951}, DeepSHAP explanations satisfy properties such as local accuracy, missingness, and consistency \citep{Lundberg2017} and is a salience method.\\

In this work, we maintain literature values for most hyperparameters of the explanation methods. Exceptions are hyperparameters of explanation methods NoiseGrad, and FusionGrad. We adjust the perturbation levels of the parameters, as discussed in \citet{Bykov2021a} to ensure at most $5\%$ loss in accuracy. All hyperparameters are presented in Table \ref{tab:XAIHP} (see Appendix \ref{app:b1}).
Additionally, both Integrated Gradients and DeepSHAP require background images as reference points to calculate the explanations (see also \citet{Lundberg2017} and Appendix \ref{app:a1}). To allow for a fair performance comparison, for both methods we sample $100$ maps containing all zero values. We note that  there are other possible reference values, e.g., natural images from training, or all-one-maps, and this choice can affect the explanation performance. Lastly, the baseline for SmoothGrad, NoiseGrad, and FusionGrad can be any local explanation method, and here we use the gradient explanations. Accordingly, gradient, SmoothGrad, NoiseGrad, and FusionGrad are sensitivity methods.

\section{Evaluation techniques}\label{sec:quantus}
Due to the lack of a ground truth explanation, XAI research developed alternative metrics to assess the reliability of an explanation method. These evaluation metrics analyze different properties an explanation method should fulfill and can serve to evaluate different explanation methods \citep{Hoffman2018, Barredo_Arrieta_2020, Mohseni2021, Hedstroem22}. Following \citet{Hedstroem22}, we describe five different evaluation properties and based on the classification task from \citet{Labe_2021} we illustrate each property in a schematic diagram (See Figures \ref{fig:robustness}-\ref{fig:complexity}). 

\subsection{Robustness}

\begin{figure}[t!]
        \centering
        \includegraphics[width=0.85\textwidth]{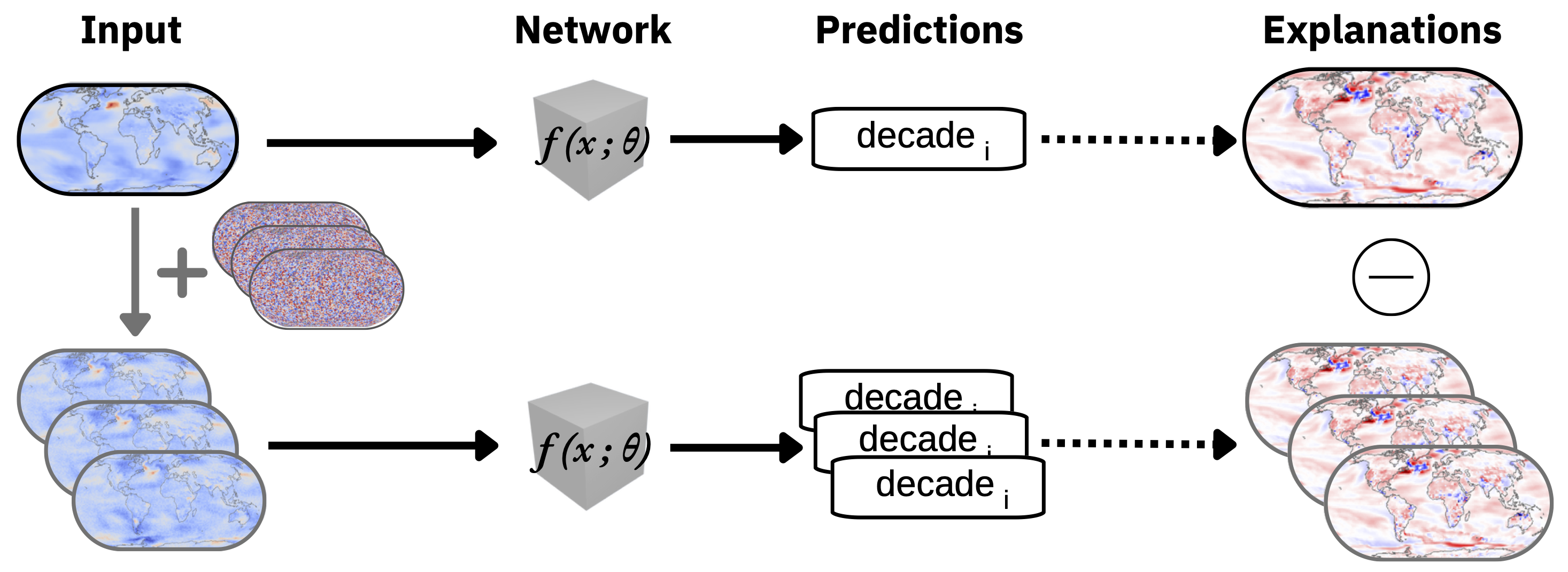}
        \caption{Diagram of the concept behind the \textit{robustness} property. Perturbed input images are created by adding uniform noise maps of small magnitude to the original temperature map  (left part of Figure). The perturbed maps are passed to the network, resulting in an explanation map for each prediction. The explanation maps of the perturbed inputs (explanation maps with grey outlines) are then compared to (indicated by a minus sign) the explanation of the unperturbed input (explanation map with black outline). A robust XAI method is expected to produce similar explanations for the perturbed input and unperturbed inputs.} 
        \label{fig:robustness}
\end{figure}

Robustness measures the stability of an explanation regarding small changes in the input image $\mathbf{x}+\delta$ \citep{Melis2018,Yeh2019,Montavon_2018}. Ideally, these small changes ($\delta<\epsilon$) in the input sample should produce only small changes in the model prediction and successively only small changes in the explanation (see Figure \ref{fig:robustness}).

To measure robustness, we choose the Local Lipschitz Estimate $q_{\text{LLE},m}$ \citep{Melis2018} and the Average Sensitivity $q_{\text{AS},m}$ \citep{Yeh2019} as representative metrics. Both use Monte Carlo sampling-based approximation to measure the Lipschitz constant or the average sensitivity of an explanation. For an explanation $\Phi^m\left(f,c, \mathbf{x}\right) \in \mathbb{R}^d$ of a XAI method $m$ and a given input $\mathbf{x}$, the scores are defined by:
\begin{equation}
    q_{\text{LLE},m} = \underset{\mathbf{x}+\delta \in \mathcal{N}_\epsilon\left(\mathbf{x}\right) \leq \
    \epsilon}{\operatorname{max}} \frac{\left\|\Phi^m\left(f,c, \mathbf{x}\right)-\Phi^m\left(f,c, \mathbf{x}+\delta\right)\right\|_2}{\left\|\mathbf{x}-(\mathbf{x}+\delta)\right\|_2},
    \label{eq:LLE}
\end{equation}
\begin{equation}
        q_{\text{AS},m} = \underset{\mathbf{x}+\delta \in \mathcal{N}_\epsilon\left(\mathbf{x}\right) \leq \
    \epsilon}{\mathbb{E}}\left[
    \frac{\|(\Phi^m(f,c, \mathbf{x})
    -\Phi^m(f,c, \mathbf{x}+\delta))\|}{\|\mathbf{x}\|}\right],
    \label{eq:AS}
\end{equation}  
where $\epsilon$ defines the discrete, finite-sample neighborhood radius $\mathcal{N}_\epsilon$ for every input $\mathbf{x} \in \mathbf{X}$, $\mathcal{N}_\epsilon\left(\mathbf{x}\right)=\left\{\mathbf{x}+\delta \in X \mid \left\|\mathbf{x}-(\mathbf{x}+\delta)\right\| \leq \epsilon\right\}$, and $c$ denotes the true class of the input sample (for more details on intuition and calculation we also suggest the primary publications \citep{Melis2018, Yeh2019}).\\ 
The robustness metrics assess the difference between the explanation of a true and perturbed image as can be seen in Eq. \eqref{eq:LLE} and \eqref{eq:AS}. Accordingly, the lowest score represents the highest robustness. 

\subsection{Faithfulness} 
\begin{figure}[t!]
        \centering
        \includegraphics[width=0.85\textwidth]{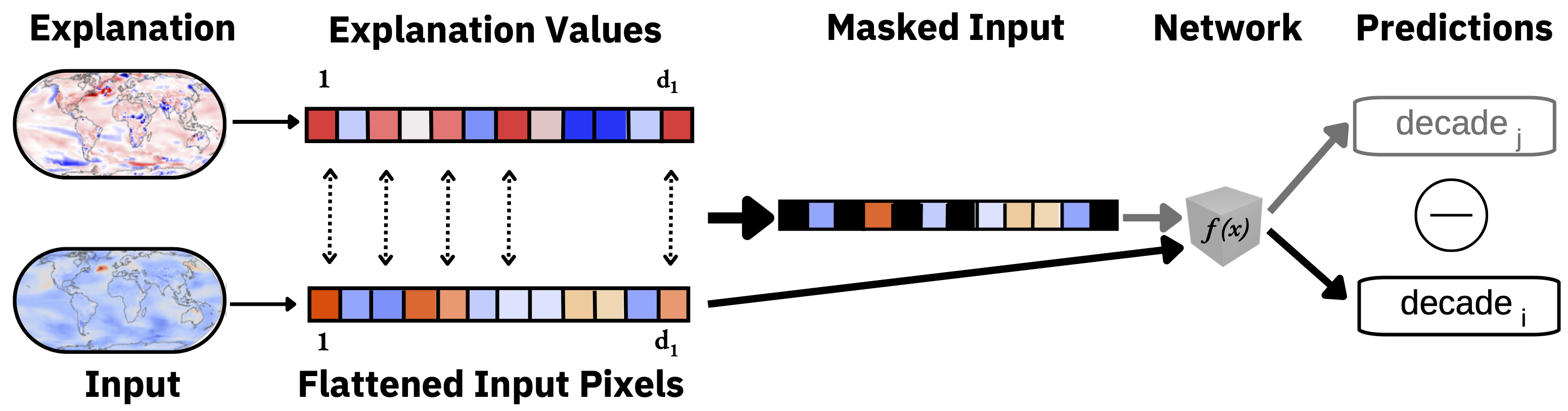}
        \caption{Diagram of the concept behind the \textit{faithfulness} property. Faithfulness assesses the impact of  highly relevant pixels in the explanation map on the network decision. First, the explanation values are sorted to identify the highest relevance values (here shown in red). Next, the corresponding pixel positions in the flattened input temperature map are identified (see dotted arrows) and masked (marked in black); i.e., their value is set to a chosen masking value, such as $0$ or $1$. Both the masked and the original input maps are passed through the network and their predictions are compared. If the masking is based on a \textit{faithful} explanation, the prediction of the masked input ($j$, grey) is expected to change compared to (indicated by a minus sign) the unmasked input ($i$, black), e.g., a different decade is predicted.} 
        \label{fig:faith}
\end{figure}

Faithfulness measures whether changing a feature that an explanation method assigned high relevance to, changes the network prediction (see Figure \ref{fig:faith}). This can be examined through the iterative perturbation of an increasing number of input pixels corresponding to high-relevance values and subsequent comparison of each resulting model prediction to the original model prediction. 
Since explanation methods assign relevance to features based on their contribution to the network's prediction, changing high-relevance features should have a larger impact on the model prediction compared to features of lesser relevance \citep{Bach2015, Samek2017, Montavon_2018, Bhatt2020, Nguyen2020}.

To measure this property, we apply RemOve And Debias (also called ROAD) \citep{Rong2022} which returns a curve of scores  $\hat{\textbf{q}} =: (\hat{q}_1,\dots,\hat{q}_I)$ for a chosen percentage range $\textbf{p}\in \mathbb{R}^{1\times I}$, with $I\in \mathbb{N}$ being the number of percentage steps (curve visualizations can also be found in \citet{Rong2022}). For each curve value $\hat{q}_i$, a percentage $p_i \in \textbf{p}, p_i \in [0,1]$ of the pixels in the input $\mathbf{x}_n$ is perturbed, according to their value in the explanation $\Phi^m\left(f,c_n, \mathbf{x}_n\right)$ (starting with the highest relevance). The predictions based on the input $\mathbf{x}_n$ and corresponding perturbed input $\hat{\mathbf{x}}_n^{i}$ are compared, resulting in $1$ for equal predictions and $0$ otherwise. The procedure is repeated for several inputs $n$. Accordingly, the ROAD score $\hat{q}^m_{ROAD, \, i}$ for each percentage $i$ corresponds to the average and is defined as:
\begin{equation}
    \hat{q}_{\text{ROAD}, m,\, i} = \frac{1}{N}\sum_{n=1}^{N}\mathbf{1}_{c_n}(c_{pred,n})\quad\text{with}\quad \mathbf{1}_{c_n}(c_{pred,n})= \begin{cases} 1 & c_n = c_{pred,n} \\ 0 & \text{otherwise} \end{cases}
    \label{eq:ROAD}
\end{equation}
where $\mathbf{1}_{c_n}:\mathbf{C}\rightarrow [0,1]$ is an indicator function comparing the predicted class $c_{pred,n}= f(\hat{\mathbf{x}}_n^i)$ of $\hat{\mathbf{x}}_n^i$ to $c_n = f(\mathbf{x}_n)$ the predicted class of the unperturbed input $\mathbf{x}_n$.
We calculate the score values for up to 50 $\%$ of pixel replacements $\mathbf{p}$ of the highest relevant pixel, calculated in steps of $1\%$; resulting in a curve $\hat{\mathbf{q}}^m_{ROAD}$. For faithful explanations, this curve should degrade faster towards increasing percentages of perturbed pixels (see Eq. \eqref{eq:Road}). The area under the curve (AUC) is then used as the final ROAD score ${q}^{ROAD,m}$:
\begin{equation}
    {q}_{\text{ROAD},m} = \text{AUC}(\mathbf{p},\hat{\mathbf{q}}^m_{ROAD})
    \label{eq:Road}
\end{equation}
Accordingly, a lower ROAD score  corresponds to higher faithfulness.\\
Furthermore, to measure faithfulness, we consider the Faithfulness Correlation $q_{\text{FC},m}$ \citep{Bhatt2020}, defined as:
\begin{equation}
    q_{\text{FC},m} = \underset{S \in |S|\subseteq d}{\operatorname{corr}}\left(\bar{\phi}^m_{S} \ , \ {f}(\mathbf{x})-{f}\left(\mathbf{x}_{\left[\mathbf{x}_{s}=\overline{\mathbf{x}}_{s}\right]}\right)\right)
    \label{eq:FC}
\end{equation}
where $S \in |S|\subseteq d$ is a set of $|S|$ random indices drawn from all pixel indices $d$ in sample $\mathbf{x}$ and $\bar{\phi}^m_{S}:=\sum_{i \in S} {\Phi}^m_{i}(f,c, \mathbf{x})$ is the sum across explanation map values $i$ that are part of the random subset $i\in S$. This set of random indices $S$ is masked (i.e. perturbed) in the input $\mathbf{x}_{\left[\mathbf{x}_{S}=\overline{\mathbf{x}}_{S}\right]}$, with $\overline{\mathbf{x}} \in \mathbb{R}^d$ being an array filled with the perturbation values (e.g. $0$ or $1$), which are used to replace all indices $i$ in the perturbed input $\mathbf{x}_{\left[\mathbf{x}_{S}=\overline{\mathbf{x}}_{S}\right]}$. Accordingly, the correlation of the prediction difference between perturbed and unperturbed input ${f}(\mathbf{x})-{f}\left(\mathbf{x}_{\left[\mathbf{x}_{s}=\overline{\mathbf{x}}_{s}\right]}\right)$, and the sum across the explanation values of the perturbed pixels $\bar{\phi}^m_{S}$ is calculated (see \citet{Bhatt2020} for more details and visualizations). Unlike ROAD, the Faithfulness Correlation score increases as the faithfulness improves.

\begin{figure}[t!]
        \centering
        \includegraphics[width=0.85\textwidth]{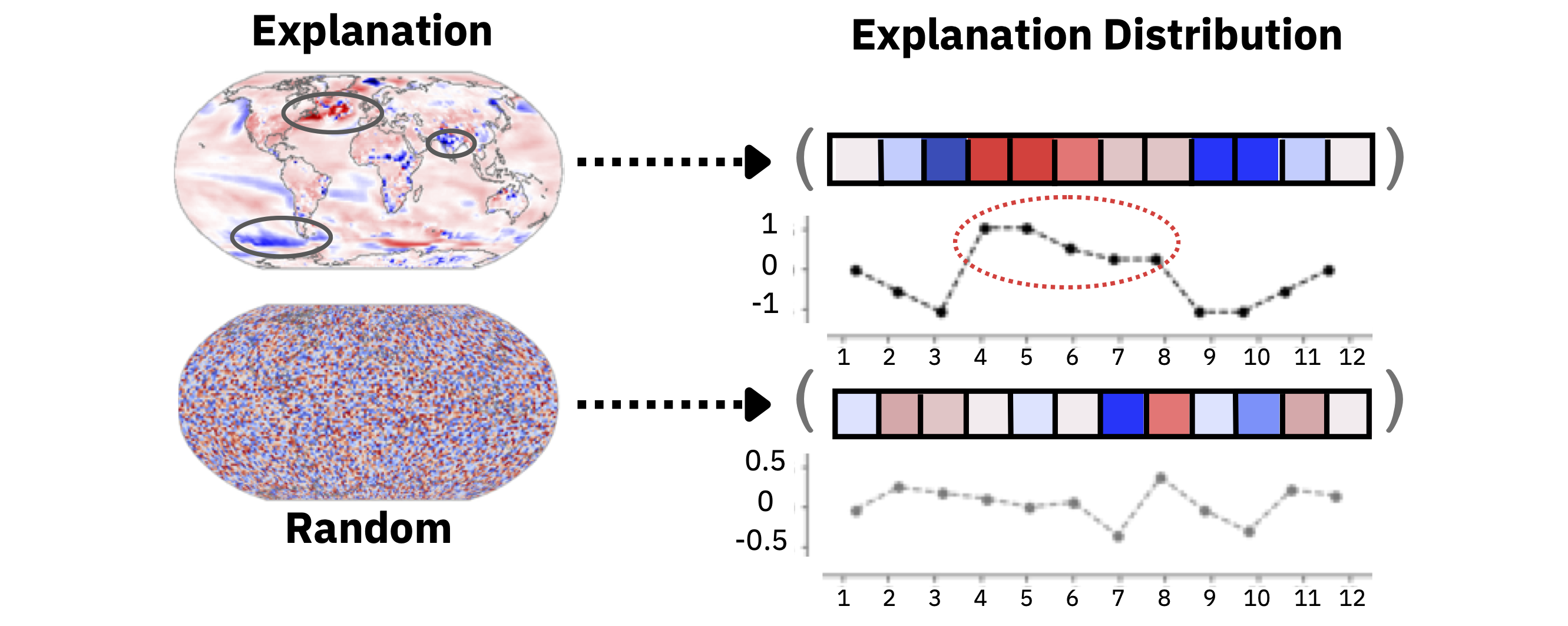}
        \caption{Diagram of the concept behind the \textit{complexity} property. Complexity assesses how the evidence values are distributed across the explanation map. For this, the distribution of the relevance values from the original explanation is compared to a 
         ``random’’ explanation drawn from a random uniform distribution. Here, shown in a 1-D example, the evidence distribution of the explanation exhibits clear maxima and minima (see maxima in red oval), which is considered desirable and linked to increased scores. The noisy features show a uniform distribution linked to a low complexity score.}
        \label{fig:complexity}
\end{figure}
\subsection{Complexity} Complexity is a measure of conciseness, indicating an explanation should consist of a few highly important features \citep{Chalasani2020,Bhatt2020} (See Figure \ref{fig:complexity}). The assumption is that concise explanations, characterized by prominent features, facilitate researcher interpretation and potentially include higher informational value with reduced noise.

Here, we use Complexity $q_{\text{COM},m}$ \citep{Bhatt2020} and Sparseness $q_{\text{SPA},m}$ \citep{Chalasani2020} as representative metric functions,
which can be formulated as follows:
\begin{equation}
    q_{\text{COM},m} = \text{H} \left(\mathcal{P}({\Phi}^m)\right), \quad \text{with} \ \mathcal{P}({\Phi}^m):= \frac{{\Phi}({f},c, \mathbf{x})}{\sum_{j \in[d]}\left|{\Phi}({f},c, \mathbf{x})_j\right|} 
    \label{eq:COM}
\end{equation}
\begin{equation}
    q_{\text{SPA},m} =  \frac{\sum_{i=1}^{d}(2 i-d-1) \Phi_m(f, \mathbf{x})}{d \sum_{i=1}^{d} \Phi(f, \mathbf{x})},
    \label{eq:SPA}
\end{equation}\\ 
where $\text{H}(\cdot)$ is the Shannon entropy, $\mathcal{P}({\Phi}^m)$ is a valid probability distribution across the fractional contribution of all features $\mathbf{x}_i$ of $\mathbf{x}$ to the total magnitude of the explanation values $\sum_{j \in[d]}\left|{\Phi}({f},c, \mathbf{x})_j\right|$, $d$ is the total number of pixels in $\mathbf{x}$, $f$ is the network function and $c$ is the explained class. Sparseness is based on the Gini index \citep{Hurley2008}, while Complexity is calculated using the entropy (see also \citet{Bhatt2020} and \citet{Chalasani2020}, where both metric functions are discussed in more detail). While the lower the entropy, the less complex the explanation, a high Gini index indicates less complexity.

\begin{figure}[t!]
        \centering
        \includegraphics[width=0.85\textwidth]{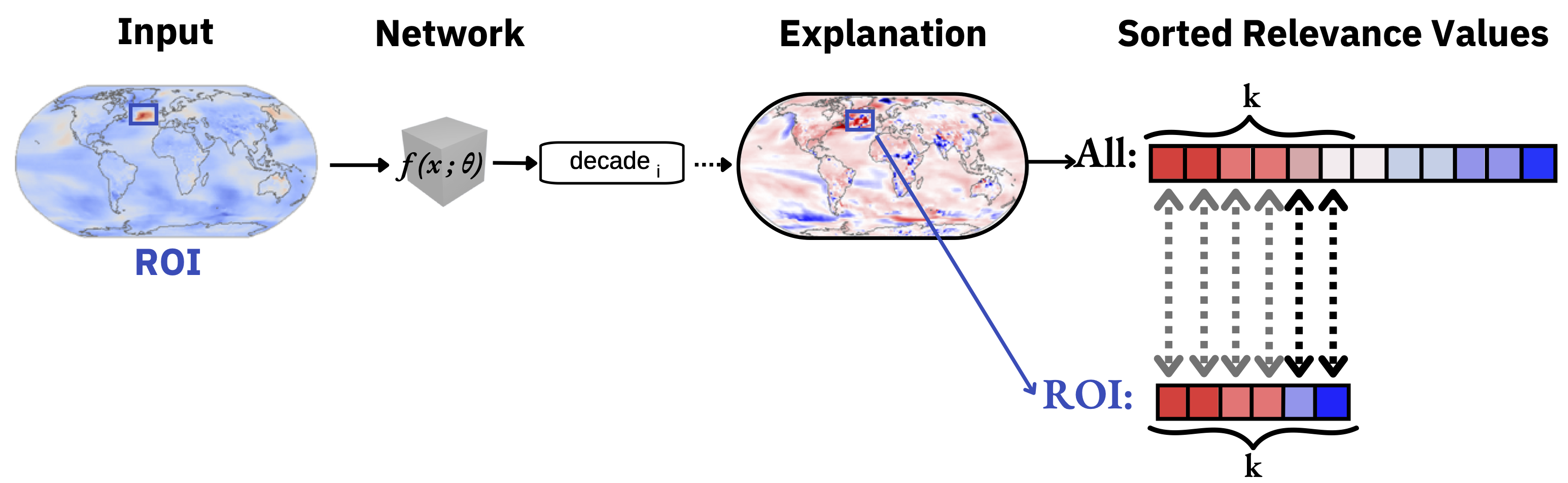}
        \caption{Diagram of the concept behind the \textit{localization} property. First, an expected region of  high relevance for the network decision, the region of interest (ROI), is defined in the input temperature map (blue box). Here, the North Atlantic is chosen, as this region has been discussed to affect the prediction (see \citet{Labe_2021}). Next, the sorted explanation values of the ROI, encompassing $k$ pixels, are compared to the $k$ highest values of the sorted explanation values across all pixels. An explanation method with strong localization should assign the highest relevance values to the ROI.}
        \label{fig:localisation}
\end{figure}
\subsection{Localization} For localization, the quality of an explanation is measured based on its agreement with a user-defined region of interest (ROI, see Figure \ref{fig:localisation}). Accordingly, the position of pixels with the highest relevance values (given by the XAI explanation) is compared to the labeled areas, e.g. bounding boxes or segmentation masks. Based on the assumption that the ROI should be mainly responsible for the network decision (ground truth) \citep{Zhang2018,Arras2020,Theiner2021,AriasDuart2021}, an explanation map yields high localization if high relevance values are assigned to the ROI.\\

As localization metrics we use the Top-$K$-pixel metric (also referred to as Top-$K$) \citep{Theiner2021} which is computed as follows:
\begin{equation}
    q_{\text{Top-$K$},m} =
    \frac{\mid{\mathbf{K}} \cap \mathbf{s}\mid}{\mid \mathbf{K} \mid},
    \label{eq:topk}
\end{equation}
where $\mathbf{K} := \Phi^m \cap \mathbf{r}_{\mid K\mid}$ denotes the vector of indices of explanation $\Phi$ corresponding to the $\mid K \mid$ highest ranked features with $\mathbf{r} = Rank\left(\Phi^m(f,c, \mathbf{x})\right)$, and $\mathbf{s}$ refers to the indices of ROI (see \citet{Theiner2021} for more details). 
Furthermore, we consider the Relevance Rank Accuracy $q_{\text{RRA},m}$ \citep{Arras2020}:
\begin{equation}
    q_{\text{RRA},m} = \frac{\mid{\Phi_\mathbf{\mid s\mid}^m} \cap \mathbf{s} \mid}{\mid \mathbf{s} \mid},
    \label{eq:RRA}
\end{equation}
where $\Phi_\mathbf{\mid s\mid}^m:= \Phi^m \cap \mathbf{r}_{\mid \mathbf{s}\mid}$ denotes the vector of indices of the explanation $\Phi$ corresponding to the highest ranked features $\mathbf{r}_{\mid \mathbf{s}\mid}\in \mathbb{R}^{1\times \mid \mathbf{s} \mid}$ and $\mid \mathbf{s} \mid$ is the number of pixels in the ROI (details on the calculation and intuition can also be found in \citet{Arras2020}). Thus, Top-$K$ and Relevance Rank Accuracy are the same for $\mid K \mid$ chosen such that it is equal to the number of pixels in the ROI $\mid \mathbf{s} \mid$. Both corresponding scores are high for well-performing methods and low for explanations with low localization.

\begin{figure}[t!]
        \centering
        \includegraphics[width=0.85\textwidth]{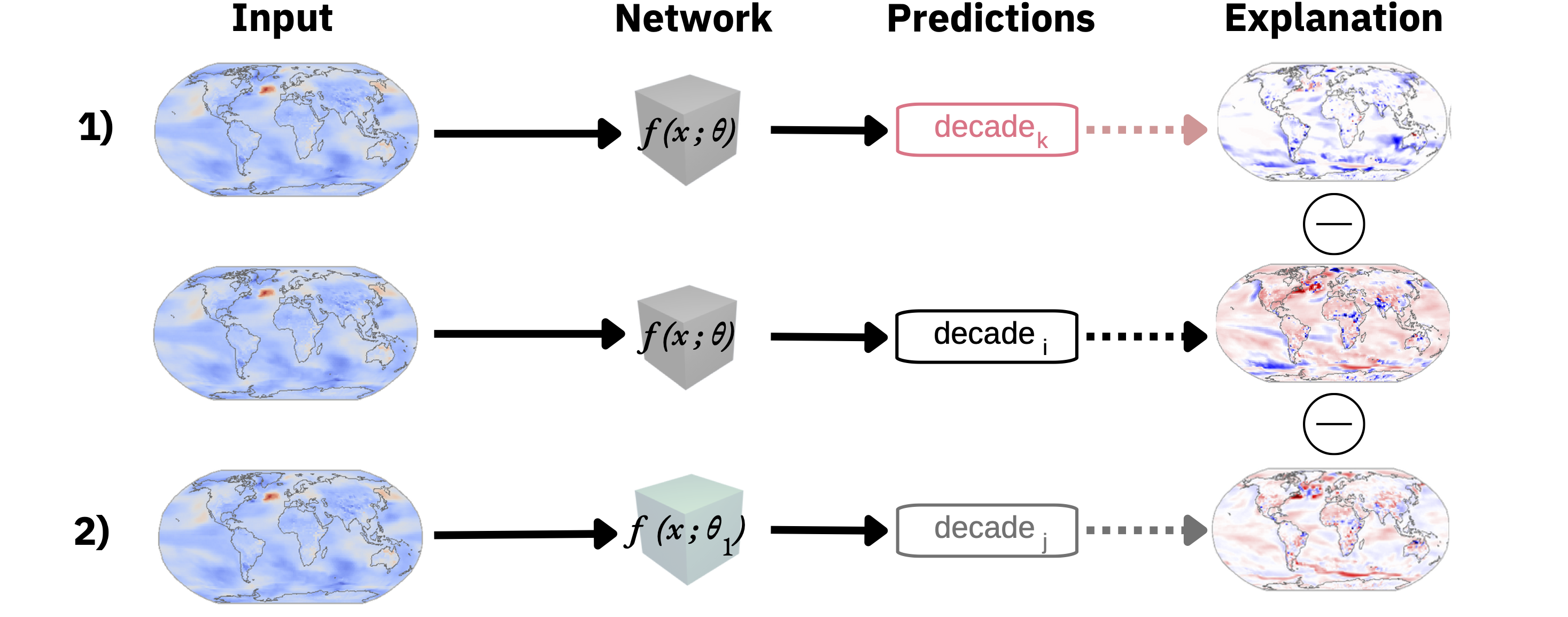}
        \caption{Diagram of the concept behind the \textit{randomization} property. In the middle row, the original input temperature map is passed through the network, and the explanation map is calculated based on the predicted (grey background) decade. For the Random Logit metric (first row - $\mathbf{1}$), the input temperature map and the network remain unchanged but the decade $k$ used to calculate the explanation is randomly chosen (pink font). The resulting explanation map is then compared to the original explanation (indicated by a minus sign) to test its dependence on the class. 
        For the Model Parametrization Randomization Test (bottom row - $\mathbf{2}$), the network is perturbed (see green box) with noisy parameters ($\theta_1=\theta + \text{noise}$), potentially altering the predicted decade ($j$, grey). The explanation map of the perturbed model should differ from the original explanation map if the explanation is sensitive to the model parameters.}
        \label{fig:randomisation}
\end{figure}
\subsection{Randomization} Randomization assesses how a random perturbation scenario changes the explanation (See Figure \ref{fig:randomisation}). Either the network weights \citep{Adebayo2018} are randomized or a random class that was not predicted by the network for the input sample $\mathbf{x}$ is explained \citep{Sixt2019}. In both cases, a change in the explanation is expected, since the explanation of an input $\mathbf{x}$ should change if the model changes or if a different class is explained.

Here, we evaluate randomization based on the Model Parameter Randomization Test \citep{Adebayo2018}. The score $q_{\text{MPT},m}$ is defined as the average correlation coefficient between the explanation of the original model $f$ and the randomized model $f_{\mathbf{W}}$ over all layers $L$:
\begin{equation}
        q_{\text{MPT},m} = \frac{1}{L} \sum_{l=1}^L \rho(\Phi^m(f,c, \mathbf{x}), \Phi^m(f_{l},c,\mathbf{x}))
        \label{eq:MPT}
\end{equation}
where $\rho$ denotes the Spearman rank correlation coefficient and $f_{l}$ is the true model with additive perturbed weights of layer $l$ (see \citet{Adebayo2018} for further details).\\
We also consider the Random Logit score $q_{\text{RL},m}$ \citep{Sixt2019}, which can be defined as e.g. structural similarity index ($SSIM$) or Pearson correlation between an explanation map of a random class $\hat{c}$ (with $f(\mathbf{x})= c$, $\hat{c}\neq c$) and an explanation map of the predicted class $c$ (see also \citet{Sixt2019} for further details and visualization):
\begin{equation}
    q_{\text{RL},m} = \text{SSIM}(\Phi^m(f,c,\mathbf{x}), \Phi^m(f,\hat{c}, \mathbf{x})).
    \label{eq:RL}
\end{equation}
Here the metrics return scores $q_{m,\, n} := q_{\mathrm{MPT/RL},\, m,\, n}$ with $n \in \{1,\dots,N\}$ for either all layers (Randomization metric) $N=L$ or all other classes ($c \neq c_{true}$) with $N = \Gamma$. Thus, we average across $L$ or $\Gamma$ to obtain $q_{m}$, as follows:
\begin{equation}
    q_{\text{MPT/RL},m}= \frac{1}{N}\sum_{n = 1}^{N}q_{m,\, n}.
    \label{eq:normRand}
\end{equation}
The metric scores of randomization and robustness metrics are interpreted similarly, i.e., low metric scores indicate strong performance.

\subsection{Metric score calculation} \label{sec:skill-baseline} 
The differing scales of the evaluation metric output (e.g. Sparseness ranges between $0-1$, Faithfulness Correlation  between $-1$ and $1$, and Local Lipschitz Estimate between $0-\infty$) and their respective interpretation (e.g. for the first two metrics the best score would be $1$, whereas for the latter the best score would be $0$) complicate their comparison. Therefore, following \citet{murphy1985}, we introduce a skill score, $S$, measuring the improvement in forecasts performance $A_f$ over the performance of reference forecast, $A_r$, relative to the perfect performance $A_p$, where $A_p=0$ if performance is measured by the mean-squared error \citep{murphy1985, murphy1988}. $S$ is given by:
\begin{equation}
    S(A_f) = \frac{A_f - A_r}{A_p - A_r}.
    \label{eq:SS}
\end{equation} 

Here, we calculate the skill score $S(q_m)$ for an explanation method based on the metric scores in each property. The skill score allows us to compare the performance of explanation methods relative to a reference score $A_r = q_r$. To establish this reference score $q_r$, we create a uniform random baseline explanation similar to \citet{Rieger2020}, maximizing the violation of each property's underlying assumptions and creating a bad-skill scenario (for details see Appendix \ref{app:a2}). The skill score then measures whether an explanation method improves upon this baseline score.

As the respective perfect score $q^\ast$ varies across metrics and takes up values of both $0$ (e.g. for Local Lipschitz Estimate) and $1$ (e.g. for Sparseness), the skill score is:
\begin{equation}
S(q_m) = \begin{cases}
1 - \frac{q_m}{q_r} & \text{if } q^\ast =0, \\ \frac{q_m - q_r}{1 - q_r} & \text{if } q^\ast = 1
\end{cases}
\label{eq:BSS-one}
\end{equation}
where $q_m \in \mathbb{R}$ represents the raw or aggregated metric score (for details see Appendix \ref{app:a2}).

\section{Experiments}
\subsection{Network predictions, explanations and motivating example}\label{sec:Motiv}
In the following, we evaluate the network performance and discuss the application of the explanation methods for both network architectures. To ensure comparability between networks and comparability to our case study \citet{Labe_2021}, we use a similar set of hyperparameters for the MLP and the CNN during training. A detailed performance discussion is provided in Appendix \ref{app:b1}. The achieved similar performance ensures that XAI evaluation score differences between the MLP and the CNN are not caused by differences in network accuracy.

After training and performance evaluation, we explain all correctly predicted temperature maps in the training, validation, and test samples (see Appendix \ref{app:b1} for details). These explanations are most often subject to further research on physical phenomena learned by the network \citep{Barnes2020,Labe_2021,Barnes2021,Labe2022}. We apply all XAI methods presented in Section \ref{sec:XAI} to both networks with the exception of the composite rule of LRP, converging to the LRP-$z$ rule for the MLP model due to its dense layer architecture \citep{Montavon2019}. The corresponding explanation maps across all XAI methods and for both networks are displayed in Figures \ref{fig:XAImlp} and \ref{fig:XAIcnn}. Despite explaining the same network predictions, different methods assign different relevance values to the same areas, revealing the disagreement problem in XAI \citep{krishna2022}. 
\begin{figure}[t!]
        \centering
        \includegraphics[width=\textwidth]{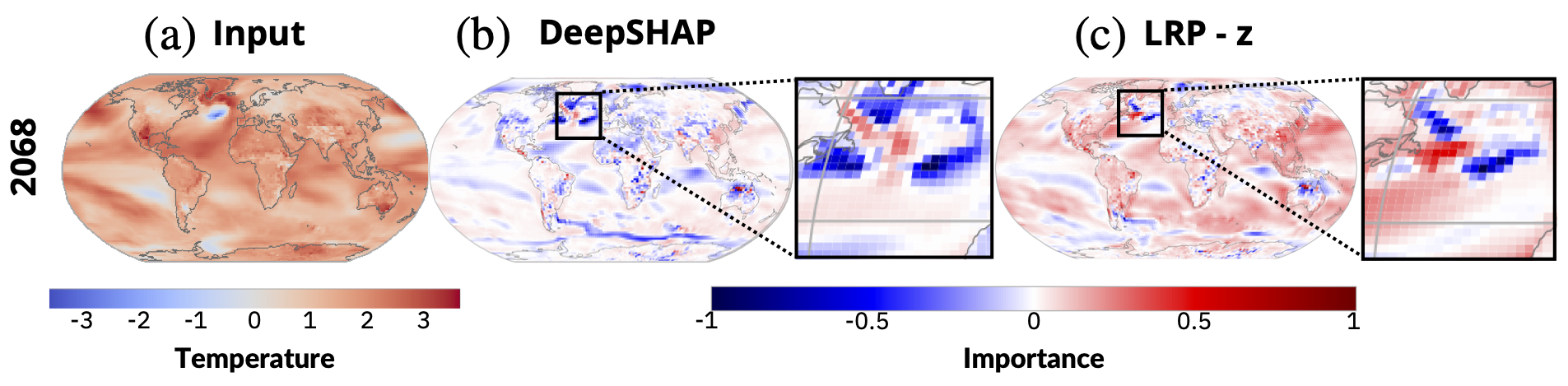}
        \caption{Motivating example visualizing the difference between different XAI methods. Shown are the T2m-temperature map (a) for the year $2068$ with the corresponding DeepSHAP (b) and LRP-$z$ (c) explanation maps of the MLP. For both XAI methods, red indicates a pixel contributed positively, and blue indicates a negative contribution to the predicted class. Next to the explanation maps, a zoomed-in map of the North Atlantic region (NA, $10-80^{\circ}$W, $20-60^{\circ}$N) is shown, demonstrating different evidence for DeepSHAP and LRP-$z$.} 
        \label{fig:Differences}
\end{figure}

To illustrate this explanation disagreement, we show the explanation maps for the year 2068 given by DeepSHAP and LRP-$z$, alongside the input temperature map in Figure \ref{fig:Differences}. According to the primary publication \citet{Labe_2021}, the cooling patch in the North Atlantic (NA), depicted in the zoomed-in map sections of $10-80^{\circ}$W, $20-60^{\circ}$N of Figure \ref{fig:Differences}, significantly contributes to the network prediction for all decades. Thus, its reasonable to assume high relevance values in this region. However, the two XAI methods display contrary signs of relevance in some areas, impeding visual comparison and interpretation. The varying sign can be attributed to DeepSHAP being based on feature-removal and modified gradient backpropagation, while LRP-$z$, in contrast, being theoretically equivalent to input times gradient. Thus, the two explanations potentially display different aspects of the network decision \citep{Clare2022} and explanations can vary in sign depending on the input image (see also discussions on input shift invariance in \citet{Mamalakis_2022}). Nonetheless, we also find common features, as for example in Australia or throughout the antarctic region. Thus, a deeper understanding of explanation methods and their properties is necessary to enable an informed method choice.  

\subsection{Assessment of explanation methods}\label{sec:metrics}
\begin{figure}[h!]
        \centering
        \includegraphics[width=\textwidth]{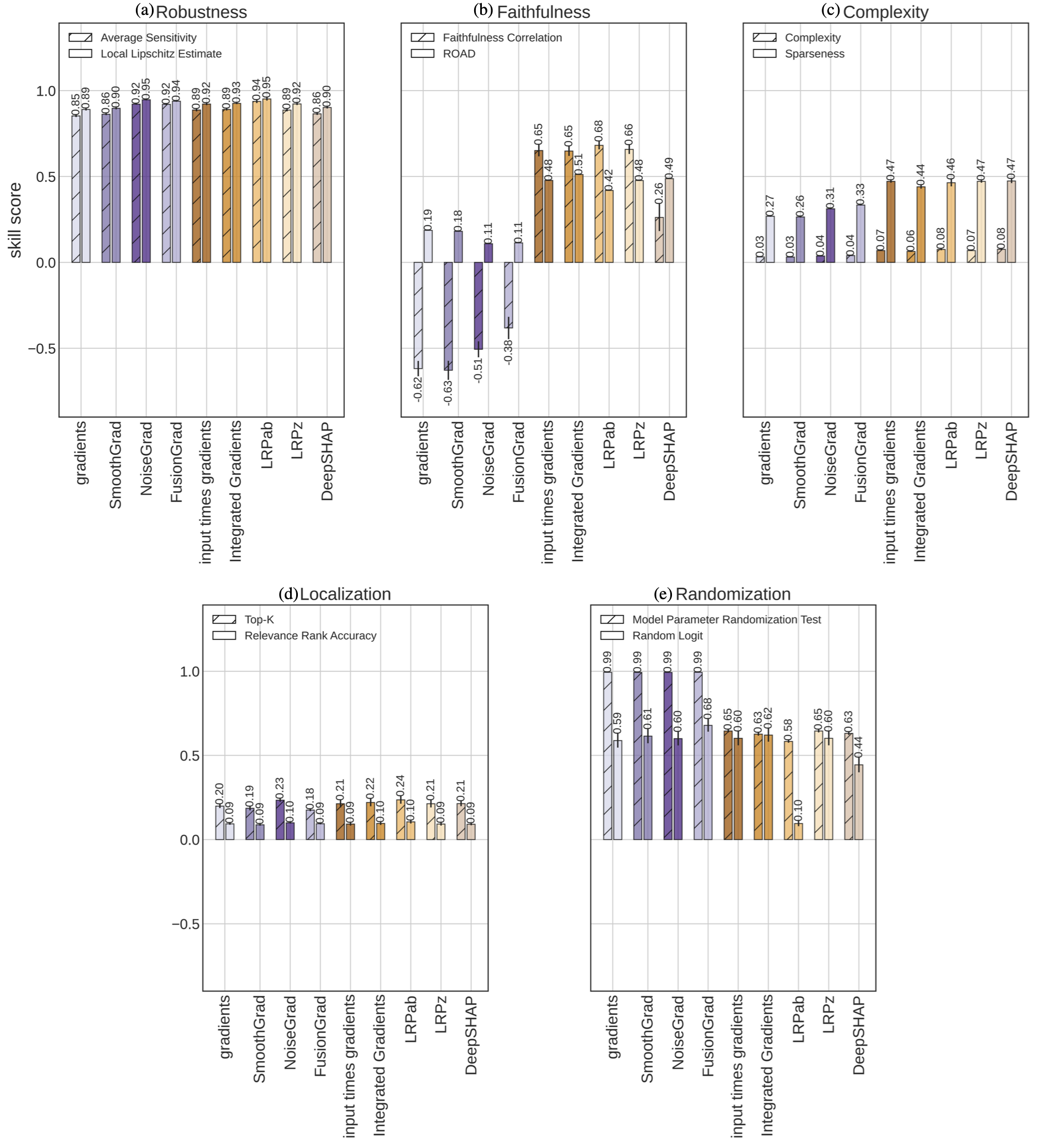}
        \caption{Barplot of skill scores based on the random baseline reference for two different metrics in each, (a) the robustness, (b) faithfulness, (c) complexity, (d) localization, and (e) randomization property. The different metrics are indicated by hatches or no hatches on the bar. We report the mean skill score (as bar labels) and the standard error of the mean (SEM), indicated by the error bars in black on each bar. The bar color scheme indicates the grouping of the XAI methods into sensitivity (violet tones) and salience/attribution methods (earthy tones).} 
\label{fig:metrics-bar}
\end{figure}

To introduce the application of XAI evaluation, we compare the different XAI methods applied to the MLP and calculate their skill scores across all five XAI method properties (see Section \ref{sec:quantus}). For each property, two representative metrics (hyperparameters are listed in Appendix \ref{app:b2}) are computed and compared. Each skill score is averaged across $50$ random samples drawn from the explanations of all correctly predicted inputs and we provide the standard error of the mean (SEM) (see Appendix \ref{app:a2} for details). To account for potential biases resulting from the choice of the testing period, we also compute the scores for random samples not limited to correct predictions. We report qualitatively robust findings (not shown) compared to the scores shown here. Our results are depicted in Figure \ref{fig:metrics-bar}.

For the \textit{robustness} property, we find that all tested explanation methods result in similar, high, and closely distributed skill scores ($\geq0.85$ and $\leq0.93$) for both the Average Sensitivity metric (hatches in Figure \ref{fig:metrics-bar}a) and Local Lipschitz Estimate metric (no hatches), where the latter shows slightly higher values overall. For both metrics, we find that salience (earthy tones) and sensitivity methods (violet tones) show a similar robustness skill and perturbation-based methods (SmoothGrad, NoiseGrad, and Integrated Gradients) do not significantly improve skill compared to the respective baseline explanations (gradient and input times gradient). We relate the latter finding to the low signal-to-noise ratio of the climate data and variability between different ensemble members, complicating the choice of an efficient perturbation threshold for the explanation methods.
Nonetheless, these findings disagree with previous studies regarding suggested robustness improvements when applying salience and perturbation-based methods \citep{smilkov2017, sundararajan2017axiomatic, Bykov2021a, Mamalakis_2022}. 

For \textit{faithfulness}, we find pronounced skill score differences between both metrics, with ROAD scores indicating positive skill for all methods, whereas Faithfulness Correlation scores include negative values for the sensitivity methods (hatched violet bars in Fig. \ref{fig:metrics-bar}b). This disparity arises from the calculation of Faithfulness Correlation metric scores using the correlation coefficient, and the distinct interpretations of relevance values in salience maps versus sensitivity maps. Since sensitivity maps display the network's sensitivity towards the change in the value of each pixel (the sign conveys the direction), the impact of the masking value depends on the discrepancy between the original pixel value and the masking value, leading to a negative correlation. 
Nonetheless, across metrics, the best skill scores $\leq 0.6$ are achieved by input times gradient, Integrated Gradients, and LRP-$z$, followed by $S(q_{\text{LPR-}\alpha-\beta}) \leq 0.42$. Furthermore, sensitivity methods (violet tones) achieve overall lower skill scores. Although DeepSHAP exhibits a lower faithfulness correlation skill (which we attribute to the challenge of applying Shapley values to continuous data \citep{han2022} and vulnerability towards feature correlation \cite{flora2022}), the method still outperforms the sensitivity methods, indicating salience (or attribution) methods provide more faithful relevance values. However, this is due to salience methods indicating the contribution of each pixel to the prediction as required by faithfulness. Thus, sensitivity methods inherently result in less faithful explanations. We note that the input multiplication of salience methods can lead to a loss of information when using standardized input pixels, as zero values in the input (i.e., values close to climatology) will result in zero values in the explanation regardless of the networks sensitivity to it (see Section \ref{sec:XAI} and \citet{Mamalakis_2022} discussing "ignorant to zero input").     


For \textit{complexity} (Figure \ref{fig:metrics-bar}c), all explanation methods exhibit low Complexity scores compared to Sparseness, indicating the explanations on climate data exhibit similar entropy to uniformly sampled values. This similarity in entropy can be attributed to the increased variability and subsequently low signal-to-noise ratio of climate data \citep{Sonnewald2021, Clare2022}. For the Sparseness metric, skill scores show skill improvement for salience (attribution) methods. We also find slight skill score improvements for NoiseGrad and FusionGrad, suggesting that incorporating network perturbations may decrease explanation complexity.

To compute the results of the \textit{localization} metrics, Top-$K$ (hatches in Fig. \ref{fig:metrics-bar}d) and Relevance Rank Accuracy (no hatches), we select the region in the North Atlantic ($10-80^{\circ}$W, $20-60^{\circ}$N) as our ROI, with the cooling in this region being a recognized feature of climate change \citet{Labe_2021}. In both metrics, all explanation methods yield low skill scores. This is consistent with lower Sparseness skill scores in complexity ($\leq 0.47$) indicating that high-relevance values are spread out, with the ROI also including fewer distinct features. In addition, high relevance in the ROI depends on whether the network learned this specific area. Thus, our results potentially indicate an inadequate choice of the ROI (either size or location) and show that localization metrics can identify a learned region. Nonetheless, LRP-$\alpha-\beta$ yields the highest skill across metrics, indicating that attributing only positive relevance values improves the distinctiveness of features in the NA region. Similar to complexity, salience methods (earthy tones) yield a slightly higher localization skill than sensitivity methods (violet tones) with the exception of NoiseGrad.

Lastly, we present the randomization results (Figure \ref{fig:metrics-bar}e). For the Random Logit metric, all XAI methods yield lower skill scores ($\geq0.1$ and $\leq 0.58$). This can be attributed to the network task classes being defined based on decades with an underlying continuous temperature trend. Thus, the differences in temperature maps can be small for subsequent years, and the network decision and explanation for different classes may include similar features. Nonetheless, we find salience (earthy tones) and sensitivity methods (violet tones) to yield no clear separation. Instead, XAI methods using perturbation result in higher skill scores, with mean improvements for FusionGrad exceeding the SEM, as well as a slight improvement for NoiseGrad and SmoothGrad over gradient and Integrated Gradients over input times gradient. Thus, while input perturbations already slightly improve the class separation in the explanation, also including network perturbation yields favorable improvement. For the Model Parameter Randomization Test scores, skill scores are overall higher ($\geq0.58$ and $\leq 0.99$) across all explanation methods, and sensitivity methods outperform salience methods, the latter aligning with \citet{Mamalakis2021a}.
Similar to the complexity results, the DeepSHAP skill score aligns with other salience method results. In addition, LRP-$\alpha$-$\beta$ yields the worst skill across metrics, potentially due to neglecting negatively contributing neurons during backpropagation (see Eq. \eqref{eq:abrule} in Appendix \ref{app:a1}) and corresponding variations across classes and under parameter randomization.

\subsection{Network-based comparison} \label{sec:Comparison}
To compare the performance of explanation methods for the MLP and CNN networks, we selected one metric per property: Local Lipschitz Estimate for robustness, ROAD for faithfulness, Sparseness for complexity, Top-$K$ for localization, and Model Parameter Randomization Test for randomization.

\begin{figure}[h!]
        \centering
        \includegraphics[width=\textwidth]{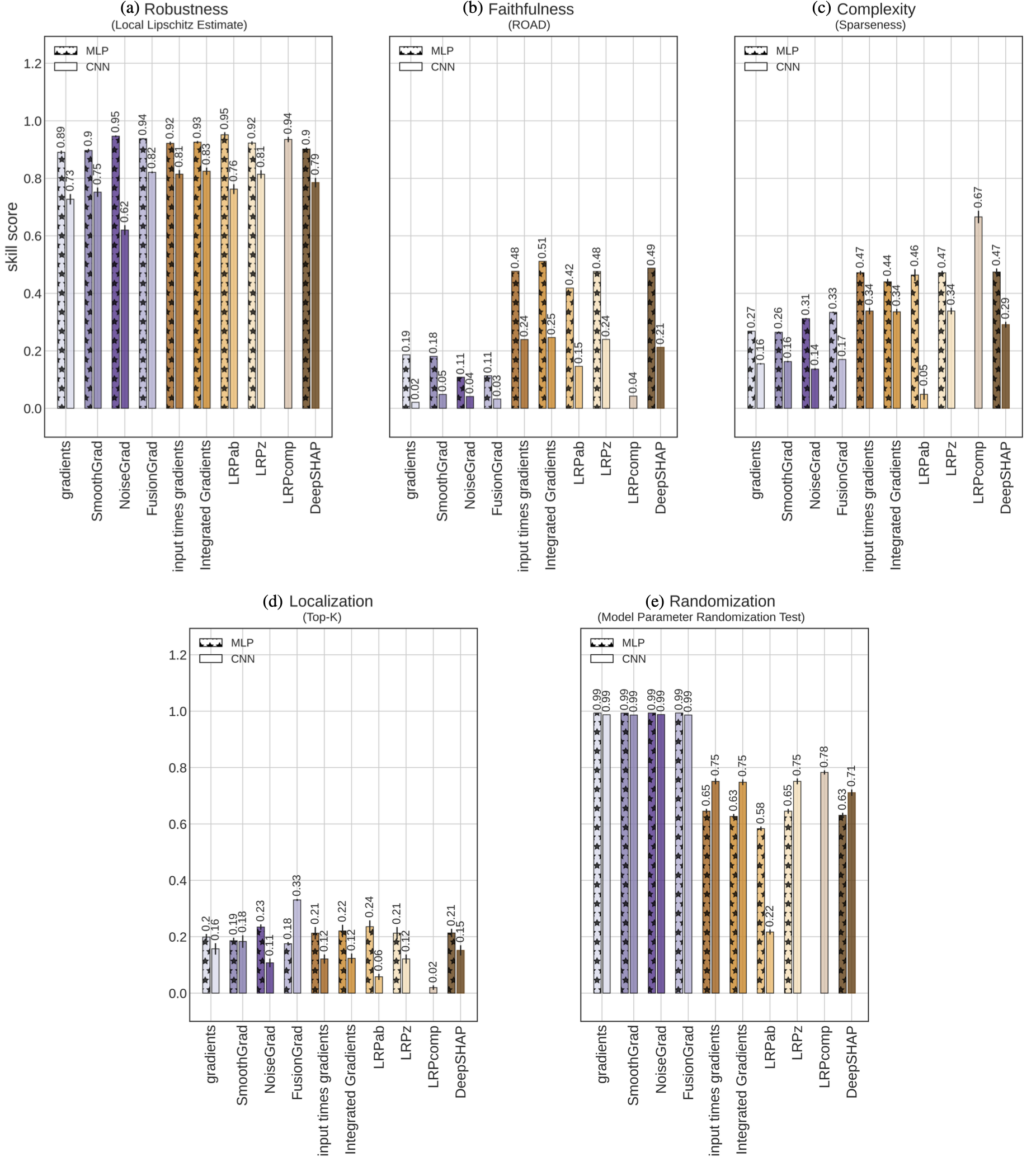}
        \caption{Barplot of skill scores based on the random baseline reference for MLP (star hatches) and CNN (no hatches) in each, the robustness (a), faithfulness (b), the complexity (c), localization (d), and randomization (e) property. We report the skill score (as bar labels) and the standard error of the mean (SEM) of all scores, indicated by the error bars in black on each bar. The bar color scheme indicates the grouping of the XAI methods into sensitivity (violet tones) and salience/attribution methods (earthy tones). Note that, for LRP-composite (LRP-$comp$) we only report the CNN results (for details, see Section \ref{sec:Motiv}).} 
\label{fig:networks-bar}
\end{figure}


For robustness (see Figure \ref{fig:networks-bar}a), XAI methods applied to the CNN yield strong skill score variations, with the MLP results showing overall higher skill scores. For the CNN, LRP-composite provides the best robustness skill. We find salience methods to exhibit slightly higher skill scores, the exception being FusionGrad outperforming LRP-$\alpha$-$\beta$ and DeepShap. This suggests that due to the differences in learned patterns between CNN and MLP, including both network and input perturbations yields more robust explanations, while the combination of a removal-based technique \citep{Covert2020} with a modified gradient backpropagation \citep{Ancona2019}as in DeepSHAP and neglecting negatively contributing neurons as in LRP-$\alpha$-$\beta$ worsens robustness compared to other salience methods. Moreover, explanation methods using input perturbations improve sensitivity explanation robustness for the CNN (SmoothGrad and FusionGrad), while methods using only network perturbations decrease robustness skill (NoiseGrad). 


In the faithfulness property (see Figure \ref{fig:networks-bar}b), salience explanation methods (Integrated Gradients, input times gradient, and LRP) achieve higher skill for both networks, aligning with previous research \citep{Mamalakis2021a, Mamalakis_2022} and the theoretical differences (see Section \ref{sec:metrics}). However, LRP-composite is the exception, adding additional insight to the findings of other studies \citet{Mamalakis_2022}, as LRP-composite sacrifices faithful evidence for a less complex (human-aligned \cite{Montavon2019}) and more robust explanation. Moreover, perturbation-based explanation methods (SmoothGrad, NoiseGrad, FusionGrad, and Integrated Gradients) do not significantly increase the faithfulness skill compared to their respective baseline explanations (gradient and input times gradient), except for Integrated Gradients for the MLP. Similar to the MLP results, LRP-$\alpha$-$\beta$ acts as an outlier compared to other salience methods. For the CNN also the DeepSHAP's faithfulness skill is decreased, contradicting theoretical claims and other findings \citep{Lundberg2017, Mamalakis_2022}. Since the CNN learns more clustered patterns (groups of pixel according to the filter-based architecture), we attribute this outcome to both DeepSHAP's theoretical definitions \citep{han2022} and vulnerability towards feature correlation \citep{flora2022}, with the latter making partitionSHAP a more suitable option \citep{flora2022}. 
 
In complexity, salience methods exhibit slight skill improvement over sensitivity methods across networks, except for LRP-$\alpha$-$\beta$ for the CNN (Figure \ref{fig:networks-bar}c). This indicates that neglecting feature relevance is more influential for the CNN's explanation, leading to fewer distinct features in the explanation, while the lower DeepSHAP skill further confirms the previously discussed disadvantages of DeepSHAP for the CNN.

In localization, both MLP and CNN show similar low overall skill scores ($\leq0.33$), indicating that the size or location of the ROI was not optimally chosen for the case study. Nonetheless, the skill scores across XAI methods are in line with the complexity results, except for the worst and best skill scores. LRP-composite yields the lowest localization skill, further confirming its trade-off between faithfulness and interpretability, also in the ROI. FusionGrad provides the highest localization skill for the CNN. In contrast, LRP-$\alpha$-$\beta$ yields the highest skill for the MLP but the second lowest skill score for the CNN. The difference in results across networks for complexity and localization can be attributed to differences in learned patterns (as discussed above), affecting properties that assess the spatial distribution of evidence in the image. 

Lastly, for randomization (see Figure \ref{fig:networks-bar}e), regardless of the network sensitivity methods outperform salience methods, indicating a decreased susceptibility to changes in the network parameters. While slightly lower, the randomization skill score of DeepSHAP does agrees with other salience methods aligning with \citet{Mamalakis2021a,Mamalakis_2022}.

Overall, our results show that while explanation methods applied to different network architectures retain similar faithfulness and randomization properties, their robustness, complexity, and localization properties depend on the specific architecture. 

\subsection{Choosing a XAI method} \label{sec:choice}
Evaluation metrics enable the comparison of different explanation methods based on various properties for different network architectures, allowing us to assess their suitability for specific tasks. Here, we propose a framework to select an appropriate XAI method.

Practitioners first determine which explanation properties are essential for their network task. For instance, for physically informed networks, randomization (the Model Parameter Randomization Test) is crucial, as parameters are meaningful and explanations should respond to their successive randomization. Similarly, localization might be less important if an ROI cannot be determined beforehand.
Second, practitioners calculate evaluation scores for each selected property across various XAI methods. We suggest calculating the skill score (see Section \ref{sec:quantus}) to improve score interpretability. Third and last, the optimal XAI method for the task can be chosen based on the skill scores independently or rank of explanation method, as in previous studies \citep{Hedstroem22, Tomsett2022, Rong2022a, Brocki2022, Gevaert2022}.\\
\begin{figure}[t]
        \centering
        \includegraphics[width=\textwidth]{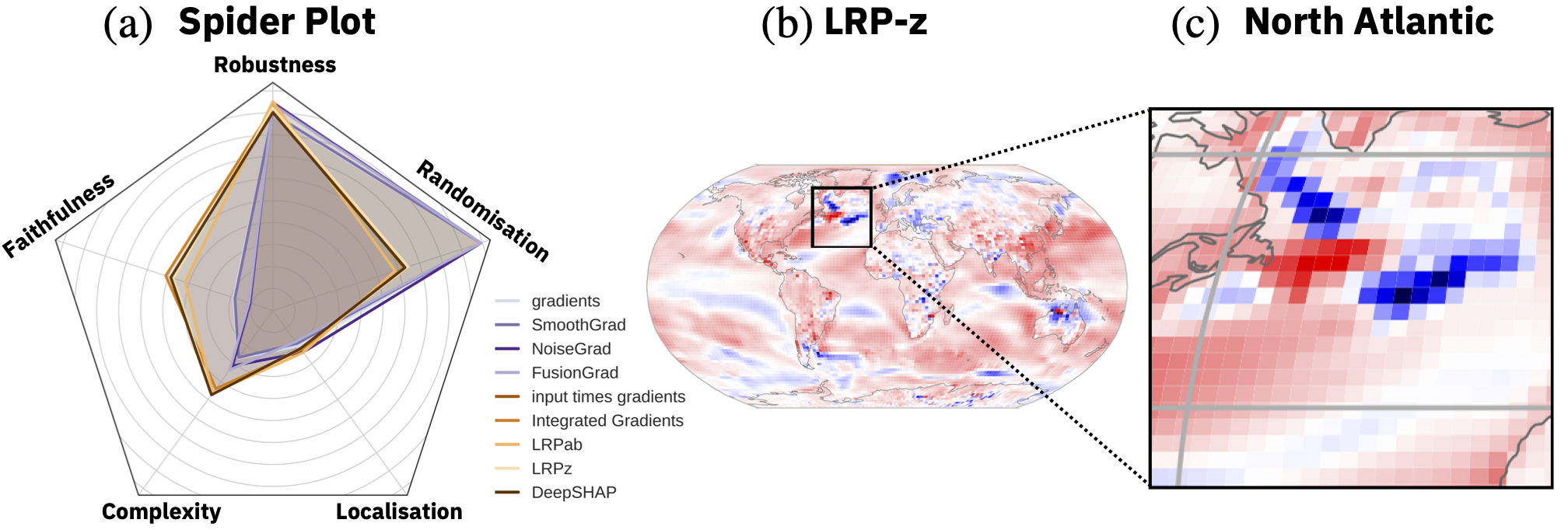}
        \caption{Visualization of the proposed procedure to choose an appropriate XAI method. In the spider plot (a) the mean skill scores for all properties across nine explanation methods (MLP explanations) are visualized, according to Figure \ref{fig:networks-bar}. The spider plot can be used as a visual aid alongside the skill scores or ranks in each essential property to identify the best-performing XAI method. In the plot, the best results correspond to the furthest distance from the center of the graph. The LRP-$z$ explanation map of the decade prediction on the temperature map of $2068$ is shown in (b) and the North Atlantic (NA) region in (c).} 
        \label{fig:spider}
\end{figure}
In our case study, for example, the explanation method should exhibit robustness towards variation across climate model ensemble members, display concise features (complexity) without sacrificing faithfulness, and capture randomization of the network parameter (randomization). Using the Quantus XAI evaluation library \citep{Hedstroem22}, we visualize the evaluation results for the MLP using a spider plot (Figure \ref{fig:spider}a), with the outermost line indicating the best-performing XAI method in each property. All methods yield similar robustness skill but differ in randomization, faithfulness and complexity skills. LRP-$z$ (light beige), input times gradient (ocher), Integrated Gradients (orange), and DeepShap (brown) provide the most faithful explanations (similar to findings in \citet{Mamalakis_2022}), with DeepShap providing a slightly worsened randomisation and robustness skill.

Based on the different strengths and weaknesses, we would select LRP-$z$ to explain the MLP predictions (Figure \ref{fig:spider}b) and analyze the impact of the NA region (Figure \ref{fig:spider}c) on the network predictions. According to the explanation, the network heavily depends on the North Atlantic region and the cooling patch pattern, suggesting its relevance in correctly predicting the decade in this global warming simulation scenario. However, we also stress that additionally applying a sensitivity method such as gradient-based SmoothGrad potentially illuminates more aspects of this network decision, as sensitivity methods provide strong randomization, in contrast to LRP-$z$.

\section{Discussion and Conclusion}\label{sec:discuss}

AI models, particularly DNNs, can learn complex relationships from data to predict unseen points afterward. However, their black box character restricts the human understanding of the learned input-output relation, making DNN predictions challenging to interpret. To illuminate the model's behavior, local XAI methods were developed, that identify the input features responsible for individual predictions and offer novel insights in climate AI research \citep{CampsValls2020, Gibson_2021, Dikshit_2021, Mayer2021, Labe_2021, van_Straaten_2022, Labe2022}. Nevertheless, the increasing number of available XAI methods and their visual disagreement \citep{krishna2022}, illustrated in our motivating example (Figure \ref{fig:Differences}), raise two important questions: Which explanation method is trustworthy, and which is the appropriate choice for a given task?

To address these questions, we introduced XAI evaluation to climate science, building upon existing climate XAI research as our case study \citep{Labe_2021}. We evaluate and compare various local explanation methods for an MLP and a CNN network regarding five properties, i.e., \textit{robustness}, \textit{faithfulness}, \textit{randomization}, \textit{complexity} and \textit{localization}, that are provided by the Quantus library \citep{Hedstroem22}. Furthermore, we improve the interpretation of the evaluation scores by calculating a skill score in reference to a random uniform explanation.

In the first experiment, we showcase the application of XAI evaluation on the MLP explanations using two metrics for each property  \citep{Melis2018,Montavon2019,Yeh2019,Bhatt2020,Arras2020,Rong2022,Hedstroem22}. Our results indicate that salience methods (i.e., input times gradient, Integrated Gradients, LRP) yield an improvement in faithfulness and complexity skill but a reduced randomization skill. Contrary to salience methods, sensitivity methods (gradient, SmoothGrad, NoiseGrad, and FusionGrad) show higher randomization skill scores while sacrificing faithfulness and complexity skills. These results indicate that a combination of explanation methods can be favourable depending on the explainability context.
We also establish that evaluating explanation methods in a climate context mandates careful consideration. For example, due to the natural variability in the data, the Sparseness metric is best suited for determining explanation complexity. Further, the Random Logit metric is favored for classification with pronounced class separations rather than datasets with continuous features spanning multiple classes. Lastly, we highlight the importance of the correct identification of an ROI to ensure an informative localization evaluation and that localization metrics enable probing the network regarding learned physical phenomena.

In the second experiment, we compare the properties of MLP and CNN explanations across all XAI methods. Both localization and complexity evaluation show larger variations between networks, due to differences in how the networks learn features in the input. The robustness results exhibit similar variation, with the CNN showing higher skill scores for all input perturbation-based methods like SmoothGrad, FusionGrad, and Integrated Gradients, contrary to the MLP, with the exception of NoiseGrad. Independent of network architecture, explanations using averages across input perturbations, like SmoothGrad and Integrated Gradients, do not consistently increase and, in some cases, even decrease the faithfulness skill. Furthermore, sensitivity methods result in less faithful and more complex explanations but capture network parameter changes more reliably. In contrast, salience methods are less complex, except for LRP-$\alpha$-$\beta$ explaining the CNN. Moreover, salience methods exhibit a higher faithfulness skill and lower randomization skill compared to sensitivity methods, consistent with findings in \citet{Mamalakis2021a, Mamalakis_2022} and in line with salience methods presenting the contribution of each input pixel rather than sensitivity (see Section \ref{sec:metrics}), due to input multiplication.
Contrary to previous research \citep{Mamalakis_2022}, LRP-$composite$ was an outlier among salience methods, sacrificing a faithful explanation for an improved complexity skill and higher robustness. Similarly, LRP-$\alpha$-$\beta$ and DeepSHAP stands out as an exception among salience methods applied to the CNN due to almost consistently lower skill scores. We attribute both findings to the mathematical definition of each method. While LRP-$composite$ is optimized towards improved interpretation resulting in less feature content, DeepSHAP is based on feature-removal and modified gradient backpropagation, and is vulnerable towards feature correlation, for CNN features and LRP-$\alpha$-$\beta$ neglecting negatively contributing neurons during backpropagation.

Lastly, we propose a framework using XAI evaluation to support the selection of an appropriate XAI method for a specific research task. The first step is to identify important XAI properties for the network and data, followed by calculating evaluation skill scores across the properties for different XAI methods. Then, the resulting skill scores across XAI methods can be ranked or compared directly to determine the best-performing method or combination of methods. In our case study, LRP-$z$ (alongside input times gradient and Integrated Gradients) yields suitable results in the MLP task, allowing the reassessment of our motivating example (Figure \ref{fig:Differences}) and the trustworthy interpretation of the NA region as a contributing input feature.

Overall, our results demonstrate the value of XAI evaluation for climate AI research. Due to their technical and theoretical differences \citep{Letzgus2021, han2022, flora2022}, the various explanation methods can reveal different aspects of the network decision and exhibit different strengths and weaknesses. Evaluation metrics allow to compare explanation methods by assessing their suitability and properties, in different explainability contexts. Next to benchmark datasets, evaluation metrics also contribute to the benchmarking of explanation methods. XAI evaluation can support researchers in the choice of an explanation method, independent of the network structure and targeted to their specific research problem.

\paragraph{Acknowledgments.}
This work was funded by the German Ministry for Education and Research through project Explaining 4.0 (ref. 01IS200551). M.K. acknowledges funding from XAIDA (European Union’s Horizon 2020 research and innovation program under grant agreement No 101003469). The authors also thank the CESM Large Ensemble Community Project \citep{Kay2015} for making the data publicly available. Support for the Twentieth Century Reanalysis Project version 3 dataset is provided by the U.S. Department of Energy, Office of Science Biological and Environmental Research (BER), the National Oceanic and Atmospheric Administration Climate Program Office, and by the NOAA Earth System Research Laboratory Physical Sciences Laboratory.

%
\paragraph*{Data availability statement.}
 Our study is based on the RPC8.5 configuration of the CESM1 Large Ensemble simulations (\url{https://www.cesm.ucar.edu/projects/community-projects/LENS/instructions.html}). The data is freely available (\url{https://www.cesm.ucar.edu/projects/community-projects/LENS/data-sets.html}). The source code for all experiments will be accessible at (\url{https://github.com/philine-bommer/Climate_X_Quantus}). All experiments and code are based on Python v3.7.6, Numpy v1.19 \citep{Harris2020}, SciPy v1.4.1 \citep{Virtanen_2020}, and colormaps provided by Matplotlib v3.2.2 \citep{Hunter_2007}. Additional Python packages used for the development of the ANN, explanation methods, and evaluation include Keras/TensorFlow \citep{Abadi2016}, iNNvestigate \citep{Alber2019} and Quantus \citep{Hedstroem22}. We implemented all explanation methods except for NoiseGrad and FusionGrad using iNNvestigate  \citep{Alber2019}. For XAI methods by \citep{Bykov2021a} and Quantus \citep{Hedstroem22} we present a Keras/TensorFlow \citep{Abadi2016} adaptation in our repository. All dataset references are provided throughout the study.



\appendix



\clearpage

\section{Additional Methodology}

\subsection{Explanations}\label{app:a1}
To provide a theoretical background we provide formulas for the different XAI methods we compare, in the following Section.
$\newline$
\textbf{Gradient}\\
The gradient method is the weak derivative $\nabla_x := \nabla f(\mathbf{x})$ of the network output $f(\mathbf{x})$ with respect to each entry of the temperature map $\mathbf{x} \in \mathbf{X}$  \citep{Baehrens2010}. 
\begin{equation}\label{gradient}
    \Phi(f(\mathbf{x})) = \nabla_x
\end{equation}
Accordingly, the raw gradient has the same dimensions as the input sample $\nabla_x, \mathbf{x} \in \mathbb{R}^D$.\\ 

\textbf{input times gradient}\\
input times gradient explanations are based on a point-wise multiplication of the impact of each temperature map entry on the network output, i.e., the weak derivative $\nabla_x$, with the value of the entry in the explained temperature map $\mathbf{x}$. All explanations are calculated as follows: 
\begin{equation}\label{inputgradient}
    \Phi(f(\mathbf{x})) = \nabla_x \mathbf{x}
\end{equation}
with $\Phi(f(\mathbf{x})),\nabla_x, \mathbf{x} \in \mathbb{R}^D$\\

\textbf{Integrated Gradients}\\
The Integrated Gradients method aggregates gradients along the straight line path from the baseline $\overline{\mathbf{x}}$ to the input temperature map $\mathbf{x}$. The relevance attribution function is defined as follows: 
\begin{equation}\label{integratedgrad}
    \Phi(f(\mathbf{x})) = (\mathbf{x} - \overline{\mathbf{x}}) \odot \int_{0}^{1} \nabla f(\overline{\mathbf{x}} + \alpha (\mathbf{x} - \overline{\mathbf{x}})) \,\text{d}\alpha,
\end{equation}

where $\odot$ denotes the element-wise product and $\alpha$ is the step-width from $\overline{\mathbf{x}}$ to $\mathbf{x}$.\\
\textbf{Layerwise Relevance Propagation (LRP)}\\
For LRP,  the relevances of each neuron $i$ in each layer $l$ are calculated based on the relevances of all connected neurons $j$ in the higher layer $l+1$ \citep{Samek2017,Montavon2017}.\\
For the \textit{$\alpha$-$\beta$}-\textbf{rule} the weighted contribution of a neuron $j$ to a neuron $i$, i.e., $z_{ij}=a_i^{(l)}w_{ij}^{(l,l+1)}$ with $a_i^{(l)}=x_i$, are separated in a positive $z^+_{ij}$ and negative $z^-_{ij}$ part. Accordingly, the propagation rule is defined by:
\begin{equation}\label{eq:abrule}
    R_i^{(l)} = \sum_j \left(\alpha \frac{z^+_{ij}}{\sum_{i} z^+_{i j}} + \beta\frac{z^-_{ij}}{\sum_{i} z^-_{i j}}\right)
\end{equation}
with $\alpha$ as the positive weight, $\beta$ as negative weight and $\alpha +\beta=1$ to maintain relevance conservation. We set $\alpha = 1$ and $\beta = 0$\\
The \textit{z}-\textbf{rule} accounts for the bounding that input images in image classification are exhibiting, by multiplying positive network weights $w_{ij}^+$ with the lowest pixel value $l_i$ in the input and the negative weights $w_{ij}^-$ by the highest input pixel value $h_i$ \citep{Montavon2017}. The relevance is calculated as follows:
\begin{equation}\label{eq:zrule}
    R_i^{(l)} = \sum_j \frac{z_{ij} - l_i w_{ij}^+ - h_i w_{ij}^+}{\sum_{i} z_{i j} - l_i w_{i j}^+ - h_i w_{i j}^+} 
\end{equation}\\
For the \textit{composite}-\textbf{rule} the relevances of the last layers with high neuron numbers are calculated based on LRP-$0$ (see \citet{Bach2015}), which we drop due to our small network. In the middle layers propagation is based on LRP-$\epsilon$, defined as:\\
\begin{equation}\label{eq:erule}
    R_i^{(l)} = \sum_j \alpha \frac{a_j(w_{ij}+ \gamma w_{ij}^+)}{\sum_{i} a_j(w_{ij}+ \gamma w_{ij}^+)} 
\end{equation}
The relevance of neurons in the layer before the input follows from LRP-$\gamma$
\begin{equation}\label{eq:grule}
    R_i^{(l)} = \sum_j \alpha \frac{z_{ij}}{\sum_{i} z_{i j}} 
\end{equation}
and the relevance of the input layer is calculated based on Eq. \ref{eq:zrule}.\\
\textbf{SmoothGrad}\\
The SmoothGrad explanations are defined as the average over the explanations of $M$ perturbed input images $\mathbf{x} + \mathbf{g}_i$ with $i = [1,\dots,M]$. 
\begin{equation}\label{eq:sg}
    \Phi(f(\mathbf{x})) = \frac{1}{M+1}\sum_{i=0}^M \Phi_{0}(f(\mathbf{x} + \mathbf{g}_i)) 
\end{equation}
The additive noise $\mathbf{g}_i \sim \mathcal{N}(0,\sigma)$ is generated using a Gaussian distribution.\\
\textbf{NoiseGrad}\\
NoiseGrad samples $N$ sets of perturbed network parameters $\hat{\bm{\theta}}_i = \bm{\eta}_i \bm{\theta}$ using multiplicative noise $\bm{\eta}_i\sim \mathcal{N}(\mathbf{1},\bm{\sigma})$. Each set of perturbed parameters $\hat{\bm{\theta}}_i$ results in a perturbed network $f_i(\mathbf{x}):=f(\mathbf{x};\hat{\bm{\theta}}_i)$, which are all explained by a baseline explanation method $\Phi_{0}(f(\mathbf{x}))$. The NoiseGrad explanation is calculated as follows:
\begin{equation}\label{eq:ng}
    \Phi(f(\mathbf{x})) = \frac{1}{N+1}\sum_{i=0}^N \Phi_{0}(f_i(\mathbf{x})) 
\end{equation}
with $f_0(\mathbf{x})=f(\mathbf{x})$ being the unperturbed network.\\
\textbf{FusionGrad}\\
For FusionGrad the NG procedure is extended by combining the SG procedure using $M$ perturbed input samples with NG calculations. Accordingly, FG can be calculated as follows:
\begin{equation}\label{eq:fg}
    \Phi(f(\mathbf{x})) = \frac{1}{M+1}\frac{1}{N+1}\sum_{j=0}^M\sum_{i=0}^N \Phi_{0}(f_i(\mathbf{x}_j)) 
\end{equation}\\
\textbf{Deep SHapley Additive exPlanations (DeepSHAP)} \citep{Lundberg2017}\\
The Deep SHAP Explainer, uses the concept of DeepLift \citep{shrikumar2016} to  approximate Shapley values. 
Formally, we can express the Shapley values as follows:\\
\begin{equation}\label{shap}
    \phi_{d_i}(f_{W}, x) = \sum_{S\subset d\setminus d_i}\frac{|S|!(|d| - |S| - 1)!}{|d|!} [f(x) - f(x_S)],
\end{equation}
where $x$ is the input with features $d$ and individual features $d_i\in d$, $f$ is our model and $x_S := x \setminus d_i$ is the masked input, only containing the features in $S \subset d \setminus \{d_i\}$, all subsets that do not contain the feature $d_i$.
For DeepSHAP, the network $f$ is separated into individual components $f_i$ according to the layer structure as proposed in DeepLift. Similar to Integrated Gradients, DeepSHAP uses a reference value (here chosen as an all-zero reference image), relative to which the contributions of each feature are calculated. This is achieved by determining the  multiplicators for each layer  according to the DeepLift multiplicators and the multiplicators are back-propagated to the input layer \citep{shrikumar2016, Lundberg2017}.\\

For visualizations, as depicted in Figure \ref{fig:XAImlp} and \ref{fig:XAIcnn} we maintain comparability of the relevance maps $\Phi(f(\mathbf{X}_{i,t}))=\bar{R}^{(i,t)} \in \mathbb{R}^{v x h}$ across different methods, by applying a \textit{min-max normalization} to all explanations:
\begin{equation}
    \bar{R}^i = \frac{\mathbb{I}_\text{max}R^i}{\mathrm{max}\left(r_{jk}|r_{jk}\in R^i \forall j \in [1,v]\forall k\in[1,h]\right)}-\frac{\mathbb{I}_\text{min} R^i}{\mathrm{min}\left(r_{jk}|r_{jk}\in R^i \forall j \in [1,v]\forall k\in[1,h]\right)}
    \label{eq:norm}
\end{equation}
with $\mathbb{I}_\text{min}, \mathbb{I}_\text{max}\in \mathbb{R}^{vxh}$ defining corresponding minimum/maximum indicator masks, i.e., for the minimum indicator each entry $\mathrm{\mathbf{i}}_\text{min}^{(jk)} = 1, \forall r_{jk}<0$ and $\mathrm{\mathbf{i}}_\text{min}^{(jk)} = 0\forall r_{jk}\geq0$, for the maximum indicator entries are defined reversely $\mathrm{\mathbf{i}}_\text{max}^{(jk)} = 1, \forall r_{jk}\geq0$ and $\mathrm{\mathbf{i}}_\text{max}^{(jk)} = 0$ otherwise.
The normalization maps pixel-wise relevance $r_{jk} \mapsto \bar{r}_{jk}$ with $\bar{r}_{jk} \in [-1,1]$ for methods identifying positive and negative relevance and $\bar{r}_{jk} \in [0,1]$ for methods contributing only positive relevance values.\\

\subsection{Evaluation Metrics}\label{app:a2}

\paragraph{Random Baseline}
Similar to \citet{Rieger2020}, we establish a random baseline as an uninformative baseline explanation. The artificial explanation $\Phi_{\text{rand}}\in \mathbb{R}^{h\times v}$ is drawn from a Uniform distribution $\Phi_{\text{rand}}\sim U(0,1)$.
Each time a metric reapplies the explanation function, for example in the robustness metrics when the perturbed input is subject to the explanation method, we redraw each random explanation. 
The only exception for the re-explanation step is the randomization metric as it aims for a maximally different explanation. Thus, to maximally violate the metric assumptions, we fix the explanation, emulating a constant explanation for a changing network $\Phi(\mathbf{x},f_\theta) \approx \Phi(\mathbf{x},f_{\hat{\theta}})$.\\

\paragraph{Score Calculation}

As discussed in Section \ref{sec:quantus}e, we calculate the skill score according to the optimal metric outcome. Thus, skill scores reported for the Average Sensitivity, the Local Lipschitz Estimate, the ROAD, the Complexity, the Model Parameter Randomization Test, and the Random Logit metrics are calculated based on the first case of Eq. \eqref{eq:BSS-one}, while the skill scores calculation based on Faithfulness Correlation, Top-$K$, Relevance Rank Accuracy, and Sparseness scores are calculated follows the bottom case of Eq. \eqref{eq:BSS-one}.

We calculate the mean skill scores $Q^m$ and corresponding SEM reported in Figures \ref{fig:metrics-bar}$-$\ref{fig:networks-bar} based on the skill scores of $I=50$ explanation samples. We choose this number of samples to provide valid statistics, while maintaining computational efficiency, for both networks. All samples are drawn randomly from the calculated explanations (both training and test data).
For each explanation method $M$, both mean skill scores $Q^m$ and corresponding SEM are calculated as follows:
\begin{equation}
    \begin{split}
        Q_m &= \frac{1}{I}\sum_{j = 1}^{I}\bar{q}_{m,\, j}\\
        \bar{\text{s}}^m &= \frac{\text{s}}{\sqrt{I}} 
    \end{split}
    \label{eq:mean}
\end{equation}
with $\text{s}$ being the standard deviation of the normalized scores $\bar{q}^m_i$ (see Section \ref{sec:quantus}) across explanation samples.\\ 
 
An exception is the ROAD metric, as discussed in Section \ref{sec:quantus}, the curve used in the AUC calculation results from the average of $N=50$ samples. Thus, we repeat the AUC calculation for $V=10$ draws of $N=50$ samples and calculate the mean skill score and the SEM.


\section{Additional Experiments} 
\subsection{Network and Explanation} \label{app:b1}
Aside from the learning rate $l$ ($l_{\text{CNN}} = 0.001$), we maintain a similar set of the hyperparameters to \cite{Labe_2021} and use the fuzzy classification setup for the performance validation. To assess the predictions of the network for each individual input we include the network predictions for 20CRv3 Reanalysis data, i.e., observations \citep{Slivinski_2019}. We measure performance using both the $RMSE = R$ between true $\hat{y}_{\text{true}}$ and predicted year $\hat{y}$ as well as the accuracy on the test set.
\begin{figure}[t!]
    \centering
    \includegraphics[width=0.9\textwidth]{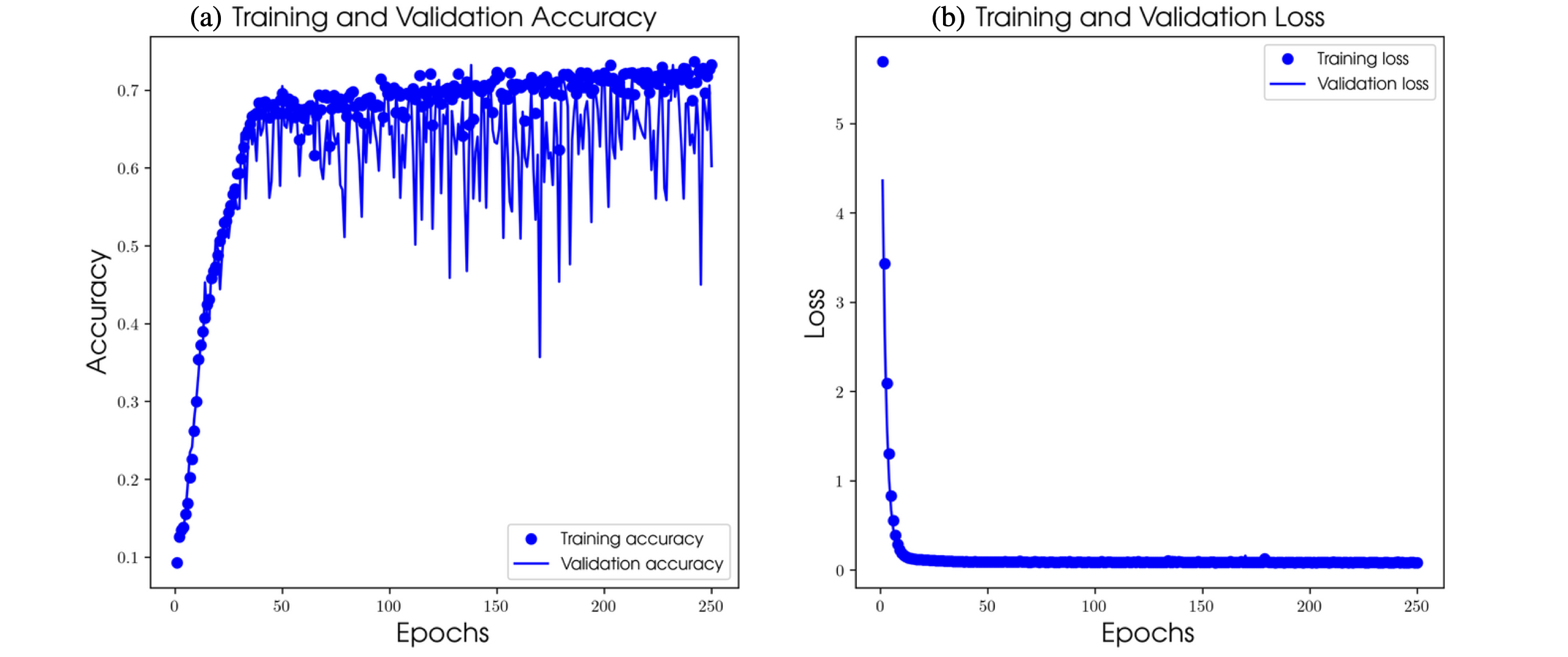}\vskip\baselineskip
    \caption{Learning curve of the MLP including accuracy (a) and loss (b). In both plots, the scatter graph represents the training performance, and the line graph the performance on the validation data.} 
    \label{fig:LearnMLP}
\end{figure}
Both the MLP and the CNN have a similar performance compared to the primary publication. We show in Figure \ref{fig:ModelResults} the regression curves for the model data (grey) and reanalysis data (blue) of A) the MLP and B) CNN (see also Figure 3c in \citet{Labe_2021}). We train both networks to exhibit no significant performance differences and prevent overfitting. 
\begin{figure}[b!]
    \centering
        \includegraphics[width=0.9\textwidth]{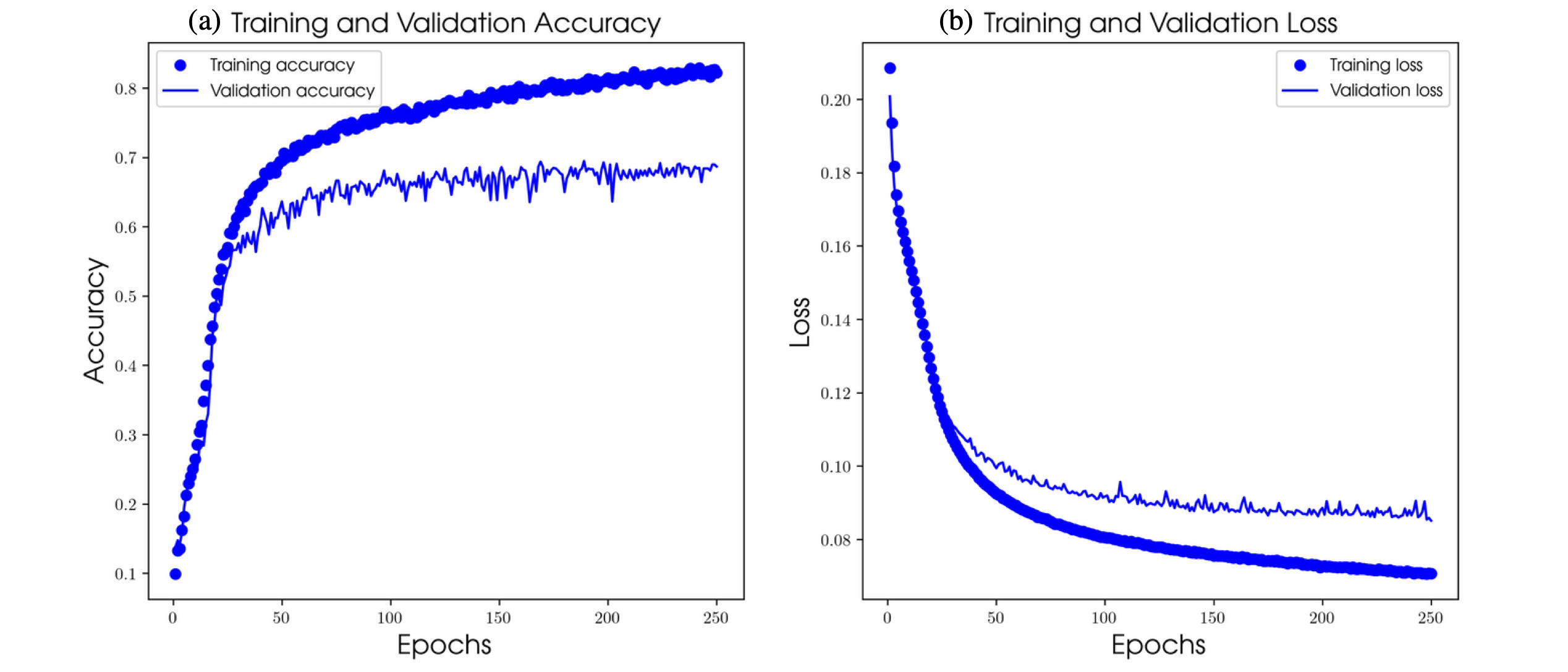}
    \vskip\baselineskip
    \caption{Learning curve of the CNN including accuracy (a) and loss (b). In both plots, the scatter graph represents the training performance, and the line graph the performance on the validation data.}
    \label{fig:LearnCNN}
\end{figure}
\begin{figure}[t!]
    \centering
        \includegraphics[width=\textwidth]{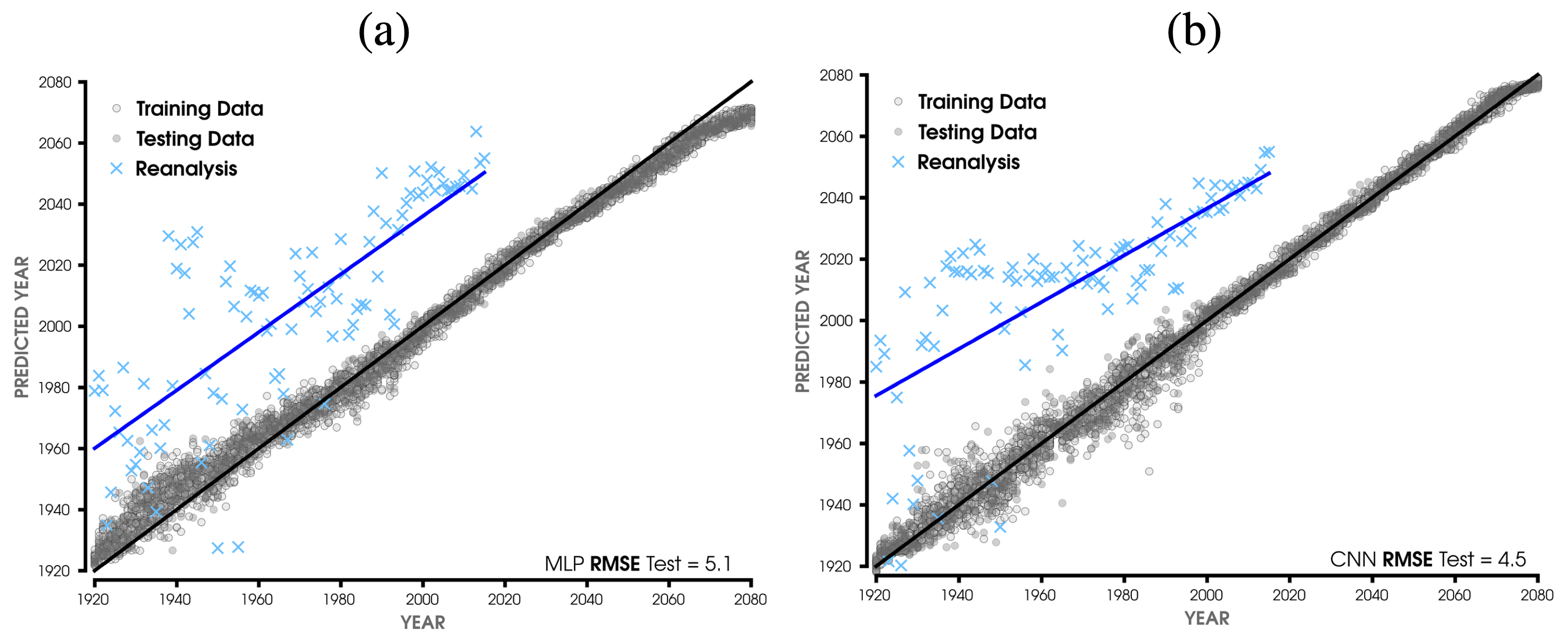}
    \vskip\baselineskip
    \caption{Network performance based on the RMSE of the predicted years to the true years of both A) MLP and B) CNN (compare to Figure 3c in \citet{Labe_2021}). The red dots correspond to the agreement of the predictions based on the training and validation data to the actual years and the grey dots show agreement between the predictions on the test set and the actual years, with the black line showing the linear regression across the full model data (training, validation and test data). In blue, we also included the predictions on the reanalysis data with the linear regression line in dark blue.} 
    \label{fig:ModelResults}
\end{figure}
The learning curves for the MLP, achieving a test accuracy of $\text{Acc}_{\text{MLP}}=67\pm4 \%$ and CNN with $\text{Acc}_{\text{CNN}}=71\pm2 \%$ (estimated across $50$ trained networks), are shown in Figures \ref{fig:LearnMLP} and \ref{fig:LearnCNN} respectively. Additionally, we consider the RMSE of the predicted years and see comparable RSME for the Test Data with $R_{\text{MLP}}=5.1$ and $R_{\text{CNN}}=4.5$.\\
In Figure \ref{fig:ModelResults} we also show the number of correct predictions for both architectures (all points on the regression line). In these graphs, we observe changing numbers of correct predictions across different years. Thus, we apply all explanation methods to the full model data $\Omega$, to ensure access to correct samples across all years.

\begin{figure}[t!]
        \centering
        \includegraphics[width=0.87\textwidth]{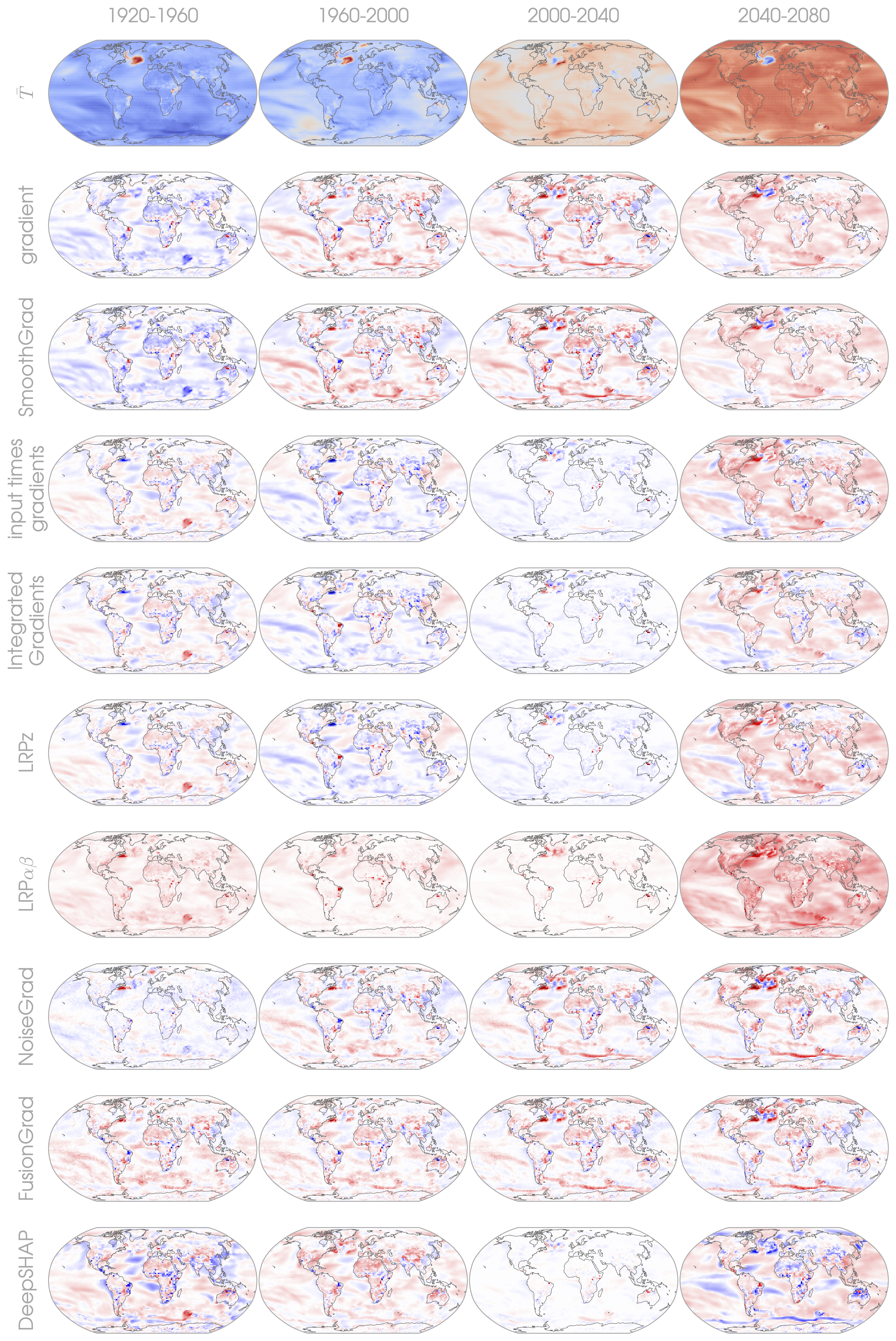}
        \caption{MLP explanation map average over $1920-1960$,$1960-2000$, $2000-2040$ and $2040-2080$ for all XAI methods. The first row shows the average input temperature map $\bar{T}$ with the color bar ranging from maximum (red) to minimum(blue) temperature anomaly. All consecutive lines show the explanation maps of the different XAI methods with the color bar ranging from $1$ (red) to $-1$ (blue).} 
        \label{fig:XAImlp}
\end{figure}
\begin{figure}[t!]
        \centering
        \includegraphics[width=0.87\textwidth]{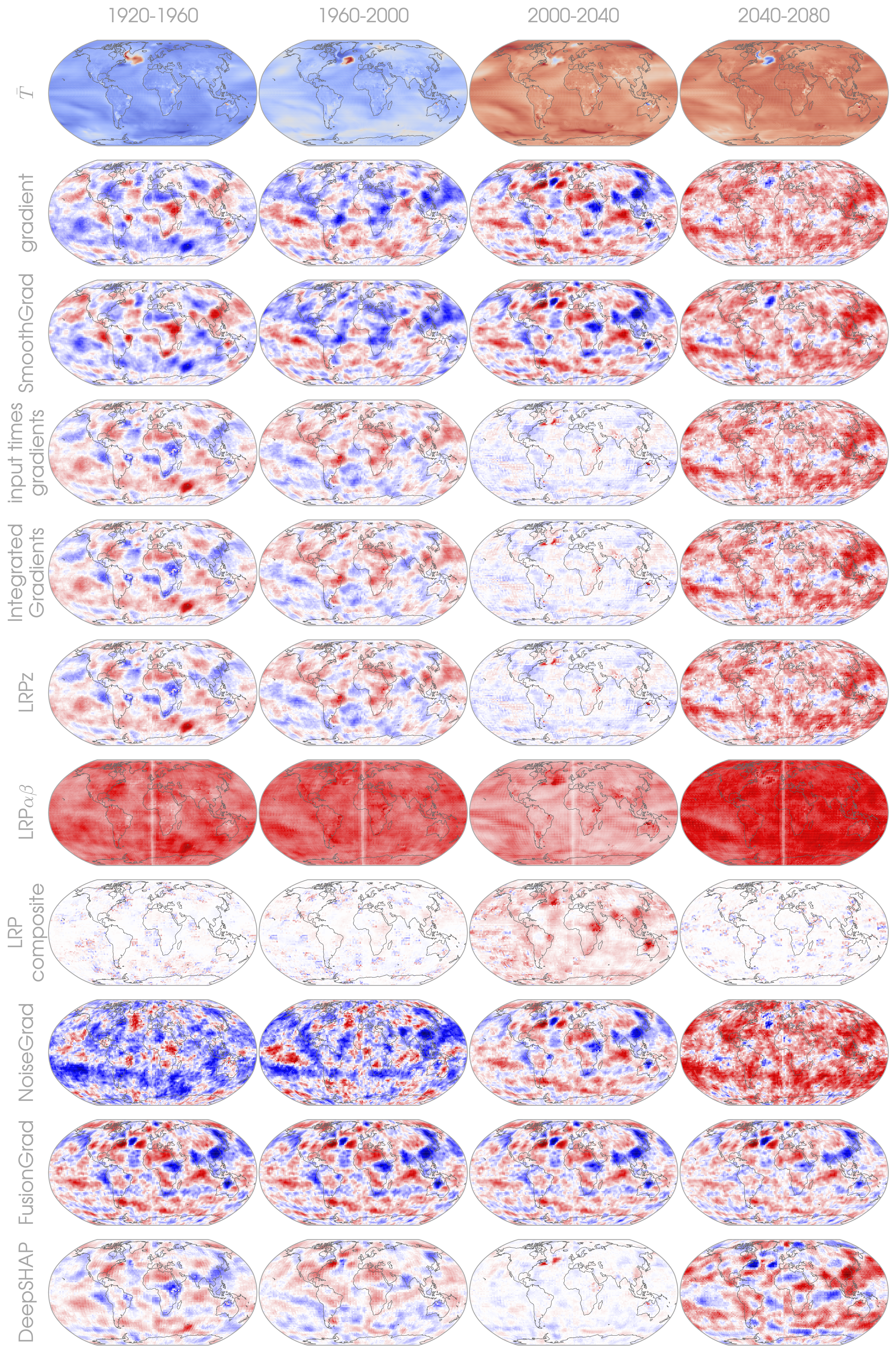}
        \caption{CNN explanation map average over $1920-1960$,$1960-2000$, $2000-2040$ and $2040-2080$ for all XAI methods. The first row shows the average input temperature map $\bar{T}$ with the color bar ranging from maximum (red) to minimum(blue) temperature anomalies. All consecutive lines show the explanation maps of the different XAI methods with the color bar ranging from $1$ (red) to $-1$ (blue).} 
        \label{fig:XAIcnn}
\end{figure}

We show examples of MLP and CNN across all explanation methods in figure \ref{fig:XAImlp} and \ref{fig:XAIcnn}. Following \citet{Labe_2021}, we adopt a criterion requiring a correct year regression within an error of $\pm 2$ years, to identify a correct prediction. We average correct predictions across ensemble members and display time periods of $40$ years based on the temporal average of explanations (see Figure 6 in \citet{Labe_2021}). \\
In comparison, both figures highlight the difference in spatial learning patterns, with the CNN relevance focusing on pixel groups whereas the MLP relevance can change pixel-wise.
In table \ref{tab:XAIHP}, we list the hyperparameters of the explanation methods, compared in our experiments. We use the notation introduced in Appendix \ref{app:a1}. We use Integrated Gradients with the baseline $\bar{\mathbf{x}}$ generated per default by iNNestigate.
\begin{table}[b!]
  \caption{
  The hyperparameters of the XAI methods. Note that parameters vary across explanation methods. We report only adjusted parameters, for all others we write $-$. We denote maximum and minimum values across all temperature maps $\mathbf{X}$ in the dataset $\Omega$ as $x_{\text{max}}$ and $x_{\text{min}}$ respectively.}
  \centering
  \resizebox{\textwidth}{!}{ 
  \begin{tabular}{l|c|c|c|c|c|c|c|c}

          {} &      \textbf{$\alpha$} & \textbf{$\beta$} & \textbf{$N$} & \textbf{$M$}& \textbf{$\sigma_{\text{SG}}$}  &     \textbf{$\sigma_{\text{NG}}$}&$\Phi_0(f(\mathbf{x}))$& $\bar{\mathbf{x}}$\\

\toprule
gradient&  $-$&$-$&$-$&$-$&$-$&$-$&$-$&$-$ \\
SmoothGrad &$-$ & $-$ &$150$ &$-$ & $0.25(x_{\text{max}}-x_{\text{min}})$& $-$& gradient&$-$ \\
NoiseGrad&  $-$ & $-$ &$-$&$20$ & $-$&$0.25$ & gradient&$-$\\
FusionGrad &  $-$ & $-$ &$20$& $20$& $0.25(x_{\text{max}}-x_{\text{min}})$& $0.125$& gradient&$-$\\
input times gradients &  $-$&$-$&$-$&$-$&$-$&$-$&$-$&$-$ \\
Integrated Gradients &  $-$&$-$&$-$&$-$&$-$&$-$&$-$&$\mathbf{0}$ \\
LRP-$\alpha$-$\beta$ &  $1$&$0$&$-$&$-$&$-$&$-$&$-$&$-$ \\
LRP-$z$ &  $-$&$-$&$-$&$-$&$-$&$-$&$-$&$-$ \\
LRP-$\text{composite}$ &  $-$&$-$&$-$&$-$&$-$&$-$&$-$&$-$ \\
DeepSHAP &  $-$&$-$&$-$&$-$&$-$&$-$&$-$&$\mathbf{0}$ \\
\bottomrule
\end{tabular}}
\label{tab:XAIHP}
\end{table}

\subsection{Evaluation metrics}\label{app:b2}
\paragraph{Hyperparmeters} In table \ref{tab:EvalHP} we list the hyperparameters of the different metrics. We list only the adapted parameters for all others (see \citet{Hedstroem22}) we used the Quantus default values. The normalization parameter refers to an explanation of normalization according to Eq. \ref{eq:norm}.\\
\textbf{Faithfulness.} In table \ref{tab:EvalHP} the perturbation function 'Indices' refers to the baseline replacement by indices of the highest value pixels in the explanation and 'Linear' refers to noisy linear imputation (see \citet{Rong2022} for details). Please, note that the evaluation of the faithfulness property strongly depends on the choice of perturbation baseline. Thus, we advise the reader to choose the uniform baseline, as determined here for standardized weather data, as it most strongly resembles noise.\\
\textbf{Randomization.} For the Model Parameter Randomization Test score calculations, we perturb the layer weights starting from the output layer to the input layer, referred to as  'bottom$\_$up' in table \ref{tab:EvalHP}. To ensure comparability we use the Pearson correlation as the similarity function for both metrics.\\
\textbf{Localisation.} For top-$k$ we consider $k=0.1 d$, which are the $10\%$ most relevant pixels of all pixels $d$ in the temperature map.
\begin{table}[t!]
  \caption{
  We show the hyperparameters of the XAI evaluation metrics based on the QUANTUS package calculations \citep{Hedstroem22}. We consider the metrics, Average Sensitivity (AS), Local Lipschitz Estimate (LLE), Faithfulness Correlation (FC), ROAD, Model Parameter Randomization Test (MPT), Random Logit (RL), Complexity (COM), Sparseness (SPA), Top-$K$ and Relevance Rank Accuracy (RRA). Note that parameters vary across metrics and we report settings only for existing parameters in each metric (for all others we write $-$).
  }
  \centering
  \resizebox{\textwidth}{!}{ 
  \begin{tabular}{l|cc|cc|cc|cc|cc}

          {} & \multicolumn{2}{c}{$\textit{Robustness}$} & \multicolumn{2}{c}{$\textit{Faithfulness}$} & \multicolumn{2}{c}{$\textit{Randomization}$} & \multicolumn{2}{c}{$\textit{Complexity}$} & \multicolumn{2}{c}{$\textit{Localisation}$}   \\
\toprule
          Hyperparameters &      $\textbf{AS}$ & $\textbf{LLE}$ & $\textbf{FC}$ & $\textbf{ROAD}$& $\textbf{MPT}$  &     $\textbf{RL}$      & $\textbf{COM}$  & $\textbf{SPA}$ & $\textbf{TopK}$  & $\textbf{RRA}$\\
\midrule
Normalization & True & True&True &True&True&True&True&True&True&True\\
Perturbation function&  $\mathcal{N}(0,0.1)$&$\mathcal{N}(0,0.1)$& Indices &Linear&$-$&$-$&$-$&$-$&$-$&$-$\\
Similarity function & Difference&\makecell{Lipschitz \\Constant} &\makecell{Pearson\\ Corr.}&$-$&\makecell{Pearson\\ Corr.}&\makecell{Pearson\\ Corr.}&$-$&$-$& $-$&$-$ \\
Num. of samples/runs  & $10$ &$10$&$50$&$-$&$-$&$-$&$-$&$-$&$-$&$-$\\
Norm nominator& Frobenius &Euclidean&$-$&$-$&$-$&$-$&$-$&$-$&$-$&$-$\\
Norm denominator & Frobenius &Euclidean&$-$&$-$&$-$&$-$&$-$&$-$&$-$ \\
Subset size &$-$ &$-$&$40$&$-$&$-$&$-$&$-$&$-$&$-$&$-$\\
Percentage range &$-$&$-$&$-$&$1-50\%$&$-$&$-$&$-$&$-$&$-$\\
$k$ & $-$&$-$&$-$&$-$&$-$&$-$&$-$&$-$&$0.1 d$&$-$\\
Perturbation baseline &$-$&$-$&$U(0,1)$&$U(0,1)$&$-$&$-$&$-$&$-$&$-$&$-$\\
Number of Classes&$-$ &$-$&$-$&$-$&$-$&$20$&$-$&$-$&$-$&$-$\\
Layer Order&$-$ &$-$&$-$&$-$& bottom$\_$up &$-$&$-$&$-$&$-$&$-$\\
\bottomrule
\end{tabular}}
\label{tab:EvalHP}
\end{table}

 \clearpage

\bibliographystyle{abbrvnat}
\bibliography{ClimateLib}


\newpage


\end{document}